%% file: main.tex
\documentclass[final]{cvpr}

\usepackage{times}
\usepackage{epsfig}
\usepackage{graphicx}
\usepackage{amsmath}
\usepackage{amssymb}
\usepackage{comment}

\usepackage{bm}
\usepackage{algorithm}
\usepackage[noend]{algpseudocode}
\usepackage{booktabs}
\usepackage{pifont}
\usepackage{multirow}
\usepackage{subcaption}
\usepackage{numprint}
\usepackage{graphbox}
\usepackage{wrapfig}
\usepackage{setspace}

\usepackage[pagebackref=true,breaklinks=true,colorlinks,bookmarks=false]{hyperref}
\usepackage[table]{xcolor}

\usepackage{color}
\input{macros}

\begin{document}

\title{Neural Geometric Level of Detail:
\\ Real-time Rendering with Implicit 3D Shapes}

\author{
    Towaki Takikawa$^{1,2,4}\thanks{Authors contributed equally.}$\hspace{0.7cm}
    Joey Litalien$^{1,3}\footnotemark[1]$  \hspace{0.75cm}
    Kangxue Yin$^{1}$  \hspace{0.75cm}
    Karsten Kreis$^{1}$ \hspace{0.75cm} 
    Charles Loop$^{1}$ \\
    Derek Nowrouzezahrai$^{3}\hspace{0.75cm}$
    Alec Jacobson$^{2}\hspace{0.75cm}$
    Morgan McGuire$^{1,3}\hspace{0.75cm}$
    Sanja Fidler$^{1,2,4}$\\
    \vspace{-3mm}\\
$^1$NVIDIA \hspace{0.75cm} $^2$University of Toronto \hspace{0.75cm} $^3$McGill University \hspace{0.75cm}  $^4$Vector Institute  \\
\vspace{-3mm}\\
\texttt{\href{https://nv-tlabs.github.io/nglod}{nv-tlabs.github.io/nglod}}
}

\maketitle

\begin{abstract}
\begingroup
\hyphenpenalty 10000
    Neural signed distance functions (SDFs) are emerging as an effective representation for 3D shapes.
    State-of-the-art methods typically encode the SDF with a
    large, fixed-size neural network to approximate complex shapes with implicit surfaces.
    Rendering with these large networks is, however, computationally expensive since it requires many forward passes
    through the network 
    for every pixel,
    making these representations impractical for real-time graphics. 
    We introduce an efficient neural representation that, for the first time, enables real-time rendering of high-fidelity neural SDFs, 
    while achieving state-of-the-art geometry reconstruction quality. 
    We represent implicit surfaces using an octree-based feature volume 
    which adaptively fits shapes with multiple discrete levels of detail (LODs), 
    and enables continuous LOD with SDF interpolation. 
    We further develop an efficient algorithm to directly render 
    our novel
    neural SDF representation in real-time 
   by querying only the necessary LODs
    with sparse octree traversal. 
    We show that our representation is 2--3 orders of magnitude more efficient
    in terms of 
    rendering speed compared to \linebreak previous works. 
    Furthermore, it produces state-of-the-art reconstruction quality for complex shapes under both 3D geometric and 2D image-space metrics. 

\endgroup
\end{abstract}

\input{sections/0_intro}

\input{sections/1_related}
\input{sections/2_method}

\input{sections/3_experiments}
\input{sections/4_conclusion}


\bibliographystyle{ieee_fullname}
{\small{\bibliography{egbib}}}


\input{sections/5_supp}

\end{document}

%% file: macros.tex
\definecolor{redcolor}{rgb}{1.0,0.0,0.0}

\input{comment_settings}
\ifnum\commentType=0
	\newcommand{\customComment}[3]{}
	\newcommand{\customTODO}[3]{}
\fi
\ifnum\commentType=1
	\newcommand{\customComment}[3]{\textcolor{#2}{\textbf{#1:} #3}}
	\newcommand{\customTODO}[3]{\textcolor{#2}{#1 TODO: #3}}
\fi
\ifnum\commentType=2
	\usepackage{pdfcomment}
	\newcommand{\customComment}[3]{\pdfcomment[icon=Comment,opacity=0.5,color=#2,author=#1]{#3}}
	\newcommand{\customTODO}[3]{\pdfcomment[icon=Note,opacity=0.5,color=#2,author=#1]{#3}}
\fi
\ifnum\commentType=3
	\usepackage{pdfcomment}
	\usepackage{todonotes}
	\usepackage{silence}
	\newcommand{\customComment}[3]{\todo[color=#2!40,size=\small]{\textbf{#1:} #3}}
	\newcommand{\customTODO}[3]{\todo[inline,color=#2!40,size=\small]{\textbf{#1:} #3}}
\fi

\newcommand{\gradient}{\nabla}
\newcommand{\concat}[2]{[{#1}\,,{#2}]}

\newcommand{\pred}[1]{\widehat{#1}}
\newcommand{\loss}{\mathcal{L}}

\newcommand{\expectation}{\mathbb{E}}
\newcommand{\real}{\mathbb{R}}
\renewcommand{\natural}{\mathbb{N}}
\newcommand{\surface}{\mathcal{S}}
\newcommand{\volume}{\mathcal{M}}
\newcommand{\faces}{F}
\newcommand{\point}{\mathbf{x}}
\newcommand{\shapeSpace}{\mathcal{Z}}

\newcommand{\sdf}{f}
\newcommand{\dist}{d}
\newcommand{\netParam}{\theta}

\newcommand{\net}{\sdf_\theta}
\newcommand{\encoding}{\psi}

\newcommand{\hiddenDim}{h}
\newcommand{\lod}{\ell}
\newcommand{\lodd}{L}

\newcommand{\maxLOD}{{\lodd_\text{max}}}
\newcommand{\shape}{\mathbf{z}}

\newcommand{\lerpParam}{\alpha}

\newcommand{\allVoxels}{\mathcal{V}}
\newcommand{\voxel}{V}
\newcommand{\voxelRes}{r}

\newcommand{\initRes}{\voxelRes_0}

\newcommand{\featureDim}{m}

\newcommand{\boundingVol}{\mathcal{B}}
\newcommand{\storage}{\mathbf{N}}
\newcommand{\decide}{\mathbf{D}}
\newcommand{\exSum}{\mathbf{S}}

\newcommand{\rays}{\mathcal{R}}
\newcommand{\ray}{\mathbf{r}}

\newcommand{\direction}{\mathbf{d}}
\newcommand{\step}{t}
\newcommand{\stIter}{k}

\newcommand{\dummy}{\,\cdot\,}
\newcommand{\eqdef}{=}
\renewcommand{\paragraph}[1]{{\vspace{1em}\noindent \textbf{#1.}}}

\renewcommand{\secref}[1]{Section~\ref{#1}}
\renewcommand{\equref}[1]{Equation~\eqref{#1}}
\renewcommand{\figref}[1]{Figure~\ref{#1}}
\renewcommand{\tabref}[1]{Table~\ref{#1}}

%% file: comment_settings.tex
\def\commentType{1}

%% file: sections/0_intro.tex
\vspace{-3mm}
\section{Introduction}
\label{sec:intro}

\input{figures/teaser}
\input{figures/moon}

Advanced geometric modeling and rendering techniques in computer graphics use 3D shapes with complex 
details, arbitrary topology, and quality, usually leveraging polygon meshes. 
However, it is non-trivial to adapt those representations to 
learning-based approaches since they lack differentiability, and thus cannot easily be used in 
computer vision 
applications such as learned image-based 3D reconstruction. Recently, neural approximations of signed 
distance functions (neural SDFs) have emerged as an attractive choice to scale up computer vision and 
graphics applications.
Prior works~\cite{Park19_DeepSDF, Mescheder_2019_CVPR, Chen_2019_CVPR, davies2020overfit} 
have shown that neural networks can encode accurate 3D geometry without restrictions on topology or resolution
by learning the SDF, which defines a surface 
by its zero level-set.
These works commonly use a large, fixed-size multi-layer perceptron (MLP) 
as the learned distance function.

Directly rendering and probing neural SDFs typically relies 
on sphere tracing~\cite{hart1996sphere}, 
a root-finding algorithm that can require hundreds of SDF evaluations {\it per pixel} to converge.
As a single forward pass through a large MLP-based SDF can require millions of operations, 
neural SDFs quickly become impractical for real-time graphics applications 
as the cost of computing a {\it single} pixel inflates to hundreds of millions of operations.
Works such as Davies et al.~\cite{davies2020overfit} circumvent this issue by using a
small neural network to overfit single shapes,
but this comes at the cost of generality and reconstruction quality. 
Previous approaches also use 
fixed-size neural networks, making them unable to express geometry with complexity exceeding
the capacity of the network.

In this paper, we present a novel representation for neural SDFs 
that can adaptively scale to different levels of detail (LODs) and 
reconstruct highly detailed geometry. Our method can
smoothly interpolate between different scales of geometry (see Figure~\ref{fig:teaser})
and can be rendered in \textit{real-time} with a reasonable memory footprint.
Similar to Davies et al.~\cite{davies2020overfit},
we also use a small MLP to make sphere tracing practical,
but without sacrificing quality or generality.

We take inspiration from 
classic surface extraction mechanisms \cite{lorensen_marching, frisken2000adaptively}
which use quadrature and spatial data structures storing distance values
to finely discretize the Euclidean space such that 
simple, linear basis functions can reconstruct the geometry. 
In such works, the resolution or tree depth determines the geometric level of detail (LOD) and 
different LODs can be blended with interpolation. 
However, they usually require high tree depths to recreate a solution with satisfying quality.

In contrast, we discretize the space by using a sparse voxel octree (SVO) and we store learned
feature vectors instead of signed distance values.
These vectors can be decoded into scalar distances using a shallow MLP, allowing us to
truncate the tree depth while inheriting 
the advantages of classic approaches (e.g., LOD). 
We additionally develop a ray traversal algorithm 
tailored to our architecture, which allows us to
render geometry close to 100$\times$ faster than DeepSDF \cite{Park19_DeepSDF}. 
Although direct comparisons with neural volumetric rendering methods are 
not possible, we report frametimes over 500$\times$ faster than NeRF \cite{mildenhall2020nerf} and 
50$\times$ faster than NSVF \cite{Liu20_NSVF} in similar experimental settings.

In summary, our contributions are as follows:\\[-5mm]
\newpage
\begin{itemize}
    \setlength\itemsep{0em}
    \item We introduce the first real-time rendering approach for complex geometry with neural SDFs. 
    \item We propose a neural SDF representation that can efficiently capture multiple LODs, and reconstruct 
          3D geometry with state-of-the-art quality (see Figure~\ref{fig:moon}).
    \item We show that our architecture can represent 3D shapes in a compressed format with higher visual fidelity than traditional methods, and generalizes across different geometries even from a single learned example.
\end{itemize}

Due to the real-time nature of our approach, 
we envision this as a modular building block for many downstream applications,
such as scene reconstruction from images, robotics navigation, and shape analysis.

%% file: figures/teaser.tex
\begin{figure}[t]
\vspace{-1.0mm}
\begin{center}
    \includegraphics[width=0.97\linewidth]{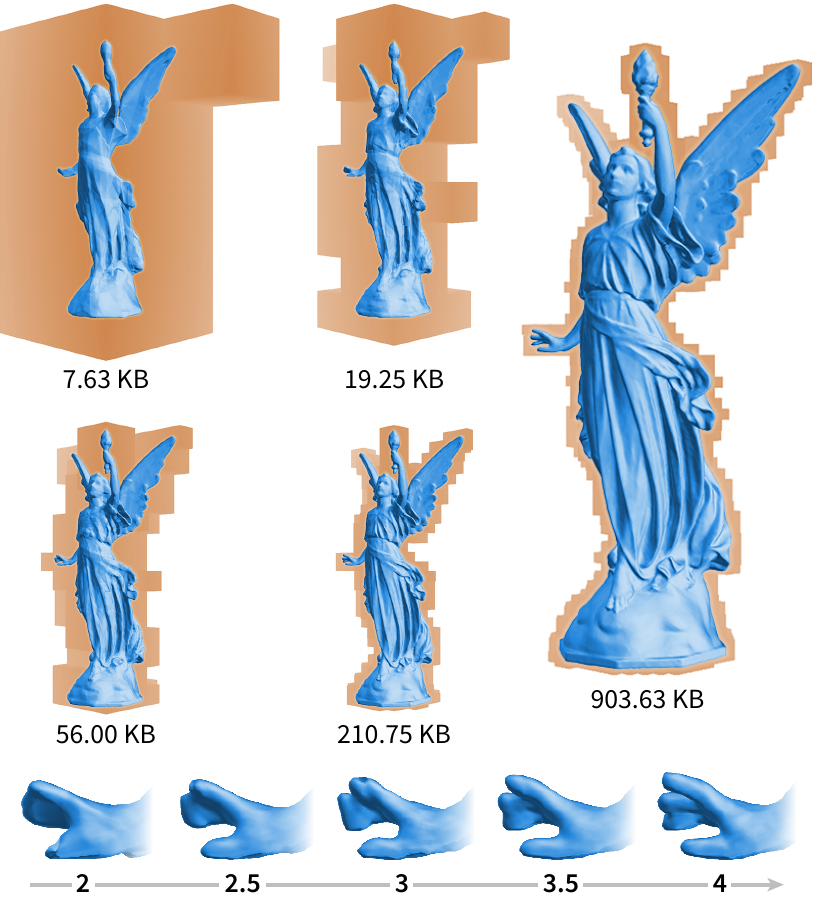}
    \vspace{-2.5mm}
      \caption{\small {\bf Levels of Detail.} 
Our representation pools features from multiple scales to 
adaptively reconstruct high-fidelity geometry with 
continuous level of detail (LOD). 
The subfigures show surfaces (blue) at varying LODs, superimposed on the 
corresponding coarse, sparse octrees (orange) which contain the features of the 
learned signed distance functions.
These were directly rendered in real-time 
using our efficient sparse sphere tracing algorithm. }
   \label{fig:teaser}
\end{center}
\vspace{-8.5mm}
\end{figure}

%% file: figures/moon.tex
\begin{figure*}[t!]
\vspace{-2.5mm}
\begin{center}
    \includegraphics[width=\linewidth]{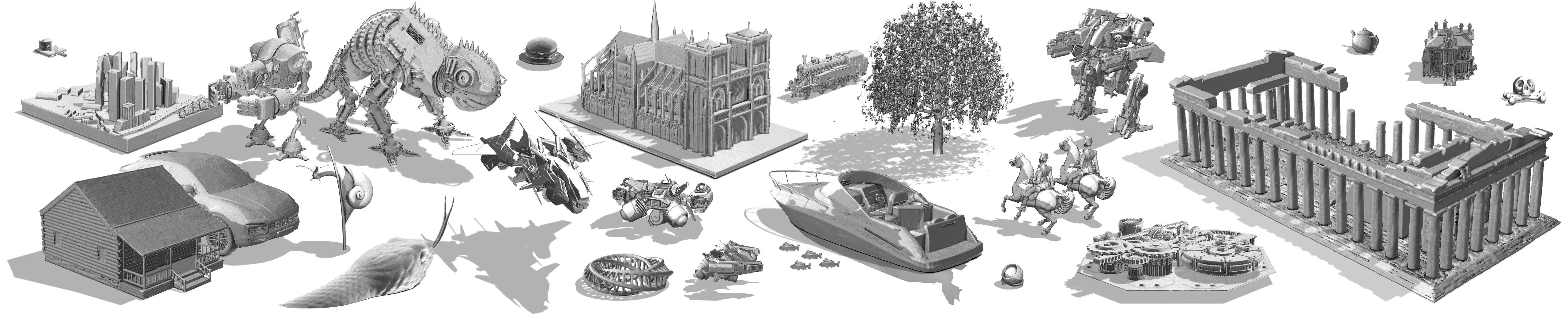}
    \vspace{-5.5mm}
    \caption{\small We are able to fit shapes of varying complexity, style, scale,
    with consistently good quality, while being able to leverage the geometry for shading,
    ambient occlusion~\cite{evans2006fast},
    and even shadows with secondary rays. {\bf Best viewed zoomed in.}
    }
\label{fig:moon}
\end{center}
\vspace{-2.0em} 
\end{figure*}

%% file: sections/1_related.tex
\section{Related Work}
\label{sec:related}

\input{figures/architecture}
Our work is most related to prior research on mesh simplification for level of detail, 
3D neural shape representations, and implicit neural rendering.

\vspace{-2mm}
\paragraph{Level of Detail}
Level of Detail (LOD)~\cite{luebke_lod} in computer graphics refers to 3D shapes 
that are filtered 
to limit feature variations, usually to approximately twice the pixel size in image space.
This mitigates flickering caused by aliasing,
and accelerates rendering by reducing model complexity.
While signal processing techniques can filter 
textures~\cite{williams1983pyramidal}, 
geometry filtering is representation-specific and challenging.
One approach is mesh decimation, 
where a mesh is simplified to a budgeted number of 
faces, vertices, or edges. Classic methods~\cite{garland_heckbert, hoppe_progressive} 
do this by greedily removing mesh elements with the  
smallest impact on geometric accuracy.
More recent methods optimize for perceptual 
metrics \cite{lindstrom_image, larkin2011perception,corsini2013perceptual}
or focus on simplifying topology~\cite{mehra2009abstraction}.
Meshes suffer from discretization errors under 
low memory constraints and have difficulty
blending between LODs. In contrast, SDFs can represent
smooth surfaces with less memory
and smoothly blend between LODs to reduce aliasing.
Our neural SDFs inherit these properties.

\vspace{-2mm}
\paragraph{Neural Implicit Surfaces} 
Implicit surface-based methods encode geometry in 
latent vectors or neural network weights, which parameterize surfaces through level-sets. 
Seminal works \cite{Park19_DeepSDF, Mescheder_2019_CVPR, Chen_2019_CVPR} 
learn these iso-surfaces by encoding the shapes into latent vectors 
using an auto-decoder---a large MLP which outputs a scalar value conditional 
on the latent vector and position.
Another concurrent line of work \cite{tancik2020fourfeat, sitzmann2019siren} uses periodic functions 
resulting in large improvements in reconstruction quality. 
Davies et al.~\cite{davies2020overfit} proposes 
to overfit neural networks to single shapes, 
allowing a compact MLP to represent the geometry.
Works like Curriculum DeepSDF \cite{duan2020curriculum} encode geometry 
in a progressively growing network,
but discard intermediate representations.
BSP-Net and CvxNet~\cite{chen2020bspnet, Deng_2020_CVPR} learn implicit geometry with space-partitioning trees.
PIFu~\cite{saito2019pifu, saito2020pifuhd} learns features on a dense 2D grid with depth as an 
additional input parameter, 
while other works learn these on sparse regular
~\cite{Genova_2020_CVPR, chabra2020deep} 
or deformed~\cite{DefTet20} 3D grids. 
PatchNets~\cite{tretschk2020patchnets} learn surface patches, 
defined by a point cloud of features. 
Most of these works rely on an iso-surface extraction algorithm like 
Marching Cubes~\cite{lorensen_marching} to create a dense surface mesh to 
render the object. In contrast, in this paper we present a method that directly renders the shape at interactive rates.

\vspace{-4mm}
\paragraph{Neural Rendering for Implicit Surfaces} 
Many works focus on rendering neural implicit representations.
Niemeyer et al.~\cite{niemeyer2020differentiable} proposes a 
differentiable renderer for implicit surfaces using ray marching.
DIST~\cite{liu2020dist} and SDFDiff~\cite{jiang2020sdfdiff} present 
differentiable renderers for SDFs using sphere tracing. 
These differentiable renderers are agnostic to the ray-tracing algorithm;
they only require the differentiability with respect to the ray-surface
intersection. As such, we can leverage the same techniques proposed in these
works to make our renderer also differentiable.
NeRF \cite{mildenhall2020nerf} learns geometry as density fields 
and uses ray marching to visualize them.
IDR \cite{yariv2020multiview} attaches an 
MLP-based shading function to a neural SDF, 
disentangling geometry and shading.
NSVF \cite{Liu20_NSVF} is similar to our work in the sense that it also 
encodes feature representations with a sparse octree. 
In contrast to NSVF, our work enables level of detail and 
uses sphere tracing, which 
allows us to separate out the geometry from shading 
and therefore optimize ray tracing,
something not possible in a volumetric rendering framework.
As mentioned previously, our renderer is two orders of magnitude faster compared
to numbers reported in NSVF \cite{Liu20_NSVF}.

%% file: figures/architecture.tex
\begin{figure*}[t]
\vspace{-3mm}
\begin{center}
    \includegraphics[width=\linewidth]{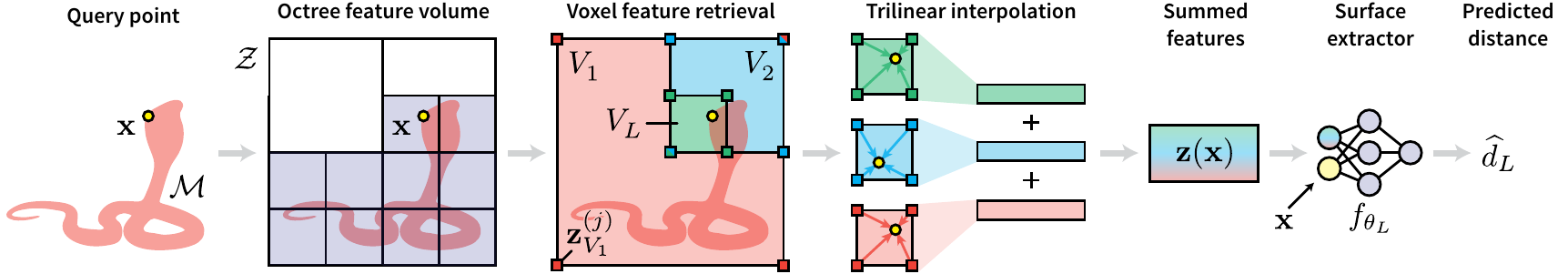}
\end{center}
\vspace{-15pt}
   \caption{\textbf{Architecture.} We encode our neural SDF using a sparse voxel octree (SVO) which
   holds a collection of features $\mathcal{Z}$.
   The levels of the SVO define LODs and the voxel corners
   contain feature vectors defining local surface segments. 
   Given query point $\point$ and LOD $\lodd$, we find corresponding voxels $\voxel_{1:\lodd}$, 
   trilinearly interpolate their corners $\shape_{\voxel}^{(j)}$ up to $\lodd$ and sum to obtain a feature vector
   $\shape(\point)$.
   Together with $\point$, this feature is fed into a small MLP $f_{\theta_\lodd}$ 
   to obtain a signed distance $\smash{\pred{\dist}_\lodd}$. 
   We jointly optimize MLP parameters $\theta$ and features $\mathcal{Z}$ end-to-end.}
\label{fig:arch}
\vspace{-6pt}
\end{figure*}

%% file: sections/2_method.tex
\section{Method}
\label{sec:method}

Our goal is to design a representation
which reconstructs detailed geometry and enables
continuous level of detail, all whilst being able
to render at interactive rates.
\figref{fig:arch} shows a visual overview of our method. 
\secref{subsec:nsdf} provides a background on
neural SDFs and its limitations. We then present our method which encodes the neural SDF in a sparse voxel octree in \secref{subsec:nglod} and provide training details in \secref{subsec:training}. Our rendering algorithm tailored to our representation is described in \secref{subsec:rendering}.

\subsection{Neural Signed Distance Functions (SDFs)}
\label{subsec:nsdf}

SDFs are functions $\sdf : \real^3 \to \real$ where $\dist = \sdf(\point)$ 
is the shortest {\it signed} distance from a point $\point$ to a 
surface $\surface=\partial\volume$ of a volume $\volume \subset \real^3$, 
where the sign indicates whether $\point$ is inside or outside of $\volume$. 
As such, $\surface$ is \textit{implicitly} represented as the 
zero level-set of $\sdf$:
\begin{equation}
    \surface \eqdef \big\{ \point \in \real^3 \,\big|\, \sdf(\point) = 0  \big\}.
\end{equation}
\begin{wrapfigure}[4]{r}{0.23\linewidth}
  \begin{center}
    \includegraphics[width=0.7in,trim=12mm 0mm 0mm 13mm]{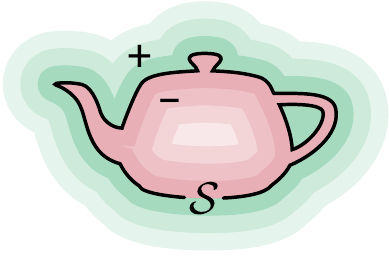}
  \end{center}
\end{wrapfigure}
A \textit{neural} SDF encodes the SDF as the parameters $\netParam$ of a neural network $\sdf_\netParam$.
\linebreak Retrieving the signed distance for a point ${\point \in\real^3}$ amounts to computing
$\sdf_\netParam(\point) = \pred{\dist}$. The parameters $\theta$ are optimized with the loss 
$J(\netParam) \eqdef \expectation_{\point,\dist}\loss\big(\net(\point), \dist \big)$, 
where $\dist$ is the ground-truth signed distance and $\loss$ is some distance metric 
such as $L^2$-distance.
An optional input ``shape" feature vector $\shape \in \real^\featureDim$ can be used 
to condition the network to fit different shapes with a fixed $\netParam$.

To render neural SDFs \textit{directly}, ray-tracing can be done with a root-finding
algorithm such as sphere tracing \cite{hart1996sphere}.
This algorithm can perform up to 
a hundred distance queries per ray, 
making standard neural SDFs prohibitively expensive 
if the network is large and the distance query is too slow. 
Using small networks can speed up this iterative rendering process,
but the reconstructed shape may be inaccurate.
Moreover, fixed-size networks are 
unable to fit highly complex shapes and cannot adapt to
simple or far-away objects where visual details are unnecessary. 

In the next section, we describe a framework that addresses these issues by
encoding the SDF using a sparse voxel octree, 
allowing the representation to adapt to different levels of
detail and to use shallow neural networks to encode geometry
whilst maintaining geometric accuracy.

\subsection{Neural Geometric Levels of Detail}
\label{subsec:nglod}

\vspace{-1em}
\paragraph{Framework}
Similar to standard neural SDFs, we represent 
SDFs using parameters of a neural 
network and an additional learned input feature which encodes the shape.
Instead of encoding shapes using a single feature vector $\shape$ as in DeepSDF~\cite{Park19_DeepSDF}, 
we use a feature volume which contains a \textit{collection} of feature vectors, which we denote by $\shapeSpace$.

We store $\shapeSpace$ in a sparse voxel octree (SVO)
spanning the bounding volume $\boundingVol = [-1,1]^3$.
Each voxel $\voxel$ in the SVO holds a learnable feature vector
$\smash{\shape_{\voxel}^{(j)}} \in \shapeSpace$ at each of its eight corners (indexed by $j$),
which are shared if neighbour voxels exist.
Voxels are allocated only if the voxel $\voxel$
contains a surface, making the SVO sparse.

Each level $\lodd \in \natural$ of the SVO defines a LOD for the geometry. As the tree depth $\lodd$
in the SVO increases, the surface is represented with finer discretization, allowing
reconstruction quality to scale with memory usage. We denote the maximum tree depth as $\maxLOD$.
We additionally employ small MLP neural networks $f_{\theta_{1:\maxLOD}}$, denoted as decoders, with parameters $\netParam_{1:\maxLOD} \eqdef \{\netParam_1,\ldots,\netParam_\maxLOD\}$ for each LOD.

To compute an SDF for a query point $\point \in \real^3$ at the desired LOD $\lodd$, we traverse the tree up to 
level $\lodd$ to find all voxels $\voxel_{1:\lodd} \eqdef \{\voxel_1,\ldots,\voxel_\lodd\}$ containing $\point$. 
For each level $\lod \in \{1,\ldots,\lodd\}$, 
we compute a per-voxel shape vector
$\encoding(\point; \lod, \shapeSpace)$ by trilinearly interpolating
the corner features of the voxels at $\point$. 
We sum the features across the levels to get $\shape(\point;\lodd,\shapeSpace)=\sum_{\lod=1}^{\lodd} \encoding(\point; \lod, \shapeSpace)$, and pass them into the MLP with LOD-specific parameters $\theta_\lodd$.
Concretely, we compute the SDF as
\begin{equation}
    \pred{\dist}_\lodd=\sdf_{\netParam_\lodd}\big(\concat{\point}{\shape(\point;\lodd,\shapeSpace)}\big),
\end{equation}
where $\concat{\dummy}{\dummy}$ denotes concatenation. 
This summation across LODs allows meaningful gradients to propagate across LODs, helping especially coarser LODs.

Since our shape vectors $\smash{\shape_{\voxel}^{(j)}}$ 
now only represent small surface segments instead of entire shapes, 
we can move the computational complexity 
out of the neural network $\net$
and into the feature vector query $\encoding : \real^3 \to \real^m$, 
which amounts to a SVO traversal and a trilinear interpolation of the voxel features.
This key design decision allows us to use very small MLPs, 
enabling significant speed-ups without sacrificing reconstruction quality.

\vspace{-1mm}
\paragraph{Level Blending} 
Although the levels of the octree
are discrete, we are able to smoothly interpolate between them.
To obtain a desired \textit{continuous} LOD $\tilde{\lodd} \geq 1$, 
we blend between different discrete octree LODs $\lodd$ by linearly interpolating the corresponding predicted distances:
\begin{equation}
    \pred{\dist}_{\tilde{\lodd}} = (1-\lerpParam)\,\pred{\dist}_{\lodd^*} + \lerpParam\,\pred{\dist}_{\lodd^*+1},
\end{equation}
where $\lodd^* = \lfloor \tilde{\lodd}\rfloor$ and $\lerpParam=\tilde{\lodd} - \lfloor \tilde{\lodd} \rfloor$ is the fractional part, allowing us to smoothly transition between
LODs (see \figref{fig:teaser}). This 
simple blending scheme only works for SDFs, and does not 
work well for density or occupancy
and is ill-defined for meshes and
point clouds.
We discuss how we set the continuous LOD $\tilde{\lodd}$ at render-time in Section \ref{subsec:rendering}.

\subsection{Training}
\label{subsec:training}
We ensure that each discrete level $\lodd$ of the SVO represents valid geometry
by jointly training each LOD. We do so by computing individual losses at each level and summing them across levels:
\begin{equation}
     J(\netParam, \shapeSpace) = \expectation_{\point,\dist}\sum_{\lodd=1}^{\maxLOD} \big\| \sdf_{\netParam_\lodd}\big(\concat{\point}{\shape(\point;\lodd,\shapeSpace)}\big) - \dist \big\|^2. \label{eq:loss}
\end{equation}
We then stochastically optimize the loss function with respect to both $\theta_{1:\maxLOD}$ and $\shapeSpace$.
The expectation is estimated with importance sampling for the points $\point\in\boundingVol$.
We use samples from a mixture of three distributions:
uniform samples in $\boundingVol$, surface samples, and perturbed surface samples. 
We detail these sampling algorithms and
specific training hyperparameters in the supplementary materials.

\subsection{Interactive Rendering}
\label{subsec:rendering}

\vspace{-1em}
\paragraph{Sphere Tracing} 
We use sphere tracing \cite{hart1996sphere} to render our representation directly. 
Rendering an SVO-based SDF using sphere tracing,
however, raises some technical implications that need to be addressed. 
Typical SDFs are defined on all of $\mathbb{R}^3$.
In contrast, our SVO SDFs are defined only for 
voxels $\voxel$ which intersect the surface geometry. 
Therefore, proper handling of distance queries made in 
empty space is required. 
One option is to use a constant step size, \ie ray marching,
but there is no guarantee the trace will converge 
because the step can overshoot.

Instead, at the beginning of the frame we first perform a ray-SVO intersection (details below) to retrieve every voxel $\voxel$ at each resolution $\lod$ that intersects with the ray. 
Formally, if ${\ray(\step) = \point_0 + \step\direction}$, $\step > 0$ is a ray with 
origin $\point_0\in\real^3$ and direction $\direction\in\real^3$, we let $\allVoxels_{\lod}(\ray)$ denote
the depth-ordered 
set of intersected voxels by $\ray$ at level $\lod$.

Each voxel in $\allVoxels_{\lod}(\ray)$ 
contains the intersected ray index, voxel position, parent voxel, and pointers to the 
eight corner feature vectors $\smash{\shape_{\voxel}^{(j)}}$.
We retrieve pointers instead of feature vectors to save memory. The feature vectors
are stored in a flatenned array, and the pointers are precalculated in an initialization
step by iterating over all voxels and finding corresponding indices to the
features in each corner. 

\input{figures/renderer.tex}

\vspace{-5pt}
\paragraph{Adaptive Ray Stepping}
For a given ray in
a sphere trace iteration $\stIter$, we perform a ray-AABB intersection \cite{Majercik2018Voxel} against 
the voxels in the target LOD level $\lodd$ to retrieve the first voxel {$\voxel_\lodd^* \in  \allVoxels_\lodd(\ray)$} that hits.
If $\point_{\stIter} \notin \voxel_\lodd^*$, we advance $\point$ to 
the ray-AABB intersection point.
If $\point_{\stIter} \in \voxel_\lodd^*$, we query our feature volume. 
We recursively retrieve all parent voxels $\voxel_\lod^*$ corresponding to the coarser levels $\lod \in \{1,...,\lodd-1\}$, 
resulting in a collection of voxels $\voxel_{1:\lodd}^*$.
We then sum the trilinearly interpolated features at each node. 
Note the parent nodes always exist by construction.
The MLP $f_{\theta_\lodd}$ then produces a 
conservative distance $\smash{\pred{\dist}_\lodd}$ 
to move in direction $\direction$, and we take a standard sphere tracing step: 
$\point_{\stIter+1} \gets \point_{\stIter} + \smash{\pred{\dist}}_\lodd \direction$.

If $\point_{\stIter+1}$ is now in empty space,
we skip to the next voxel in $\allVoxels_{\lodd}(\ray)$ along the ray and discard the ray $\ray$ if none exists.
If $\point_{\stIter+1}$ is inside a voxel, we perform a sphere trace step. This repeats
until all rays miss or if a stopping criterion is reached to recover a hit point $\point^* \in \surface$. The process is illustrated in \figref{fig:renderer}.
This adaptive stepping enables voxel sparsity by never having to query
in empty space, allowing a minimal storage for our representation. 
We detail the stopping criterion in the supplementary material.

\vspace{-1mm}
\paragraph{Sparse Ray-Octree Intersection} 
We now describe our novel ray-octree intersection algorithm that makes use of a 
breadth-first traversal strategy and parallel scan kernels~\cite{merrill2017cub}
to achieve high performance on modern graphics hardware.
Algorithm~\ref{alg:rayoctree} provides  pseudocode of our algorithm.
We provide subroutine details in the supplemental material.

\input{figures/rayoctree.tex}

This algorithm first generates a set of rays $\rays$ (indexed by $i$)
and stores them in an array $\storage^{(0)}$ of ray-voxel pairs, which are proposals for ray-voxel intersections.
We initialize each $\storage^{(0)}_i \in \storage^{(0)}$ with the root node, the octree's top-level voxel (line 2).
Next, we iterate over the octree levels $\lod$ (line 3). In each iteration, we 
determine the ray-voxel pairs that result in intersections in \textsc{Decide}, which returns 
a list of {\it decisions} $\decide$ with $\decide_j = 1$ if the ray intersects the voxel and $\decide_j = 0$ otherwise (line 4).
Then, we use \textsc{ExclusiveSum} to compute the exclusive sum $\exSum$ of list $\decide$, 
which we feed into the next two subroutines (line 5).
If we have not yet reached our desired LOD level $\lodd$, 
we use \textsc{Subdivide} to 
populate the next list $\storage^{(\lod+1)}$ with child voxels of those $\storage^{(\lod)}_j$ that the ray intersects
and continue the iteration (line 9).
Otherwise, we use \textsc{Compactify} to remove all $\storage^{(\lod)}_j$ 
that do not result in an intersection (line 7). 
The result is a compact, depth-ordered
list of ray-voxel intersections for each level of the octree.
Note that by analyzing the octant of space that the 
ray origin falls into inside the voxel, we can order the 
child voxels so that the list of ray-voxel pairs  $\storage^{(\lodd)}$ will be ordered by distance to the ray origin.

\paragraph{LOD Selection} 
We choose the LOD $\tilde{\lodd}$ for rendering with a depth
heuristic, where $\tilde{\lodd}$ transitions linearly with 
user-defined thresholds based on distance to object. 
More principled approaches exist
\cite{amanatides1984ray}, 
but we leave the details up to the user to choose an algorithm
that best suits their needs.

%% file: figures/renderer.tex
\begin{figure}[t]
\vspace{-0mm}
\begin{center}
    \includegraphics[width=1.\linewidth]{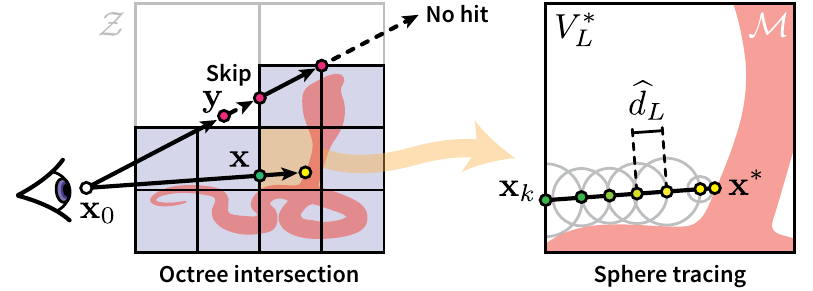}
\end{center}
\vspace{-5.0mm}
   \caption{{\bf Adaptive Ray Steps.}  When the query point is \textit{inside} a voxel (e.g., $\point$), 
   trilinear interpolation is 
   performed on all corresponding voxels up to the base octree resolution to compute a sphere tracing step (right). When
   the query point is \textit{outside} a voxel (e.g., $\mathbf{y})$, ray-AABB intersection is used to skip to the next voxel.
   \label{fig:renderer}}
\vspace{-3mm} 
\end{figure}

%% file: figures/rayoctree.tex
\begin{algorithm}[ht]
\algblockdefx{ForAll}{EndForAll}[1]%
{\textbf{for all }#1 \textbf{do in parallel}}%
{\textbf{end for}}
\algtext*{EndForAll}
\algdef{SE}[DOWHILE]{Do}{doWhile}{\algorithmicdo}[1]{\algorithmicwhile\ #1}%
\caption{Iterative, parallel, breadth-first octree traversal}\label{alg2}
\begin{algorithmic}[1]
\Procedure{RayTraceOctree}{$L, \mathcal{R}$}
    \State $\storage^{(0)}_i \gets \{i, 0\}, i=0,\ldots,|\rays|-1$
    \For{$\lod=0$ to $\lodd$}
        \State $\decide \gets \Call{Decide}{\mathcal{R}, \storage^{(\lod)}, \lod}$
        \State $\exSum \gets \Call{ExclusiveSum}{\decide}$
        \If{$\lod = \lodd$}
            \State $\storage^{(\lod)} \gets \Call{Compactify}{\storage^{(\lod)}, \decide, \exSum}$
        \Else
            \State $\storage^{(\lod+1)} \gets \Call{Subdivide}{\storage^{(\lod)}, \decide, \exSum}$
        \EndIf
    \EndFor
 \EndProcedure
\end{algorithmic}
\label{alg:rayoctree}
\end{algorithm}

%% file: sections/3_experiments.tex
\input{tables/storage}

\section{Experiments}

We perform several experiments to showcase the effectiveness of our architecture. 
We first fit our model to 3D mesh models
from datasets including ShapeNet \cite{chang2015shapenet}, 
Thingi10K \cite{zhou2016thingi10k}, and select models from TurboSquid\footnote{\url{https://www.turbosquid.com}}, and
evaluate them based on both 3D geometry-based 
metrics as well as rendered image-space metrics. 
We also demonstrate that our model is able
to fit complex analytic signed distance functions with unique properties from Shadertoy\footnote{\url{https://www.shadertoy.com}}. 
We additionally show results on real-time rendering, generalization to multiple shapes, and geometry simplification.

The MLP used in our experiments has only a single hidden layer with dimension
$\hiddenDim=128$ with a ReLU activation in the intermediate layer, thereby being
significantly smaller and faster to run than the networks used in the baselines we compare 
against, as shown in our experiments.
We use a SVO feature dimension of $\featureDim=32$.
We initialize voxel features $\shape \in\shapeSpace$ using a Gaussian prior with $\sigma=0.01$. 

\subsection{Reconstructing 3D Datasets}

We fit our architecture on several different 3D datasets, 
to evaluate the quality of the reconstructed surfaces. We compare against 
baselines including DeepSDF \cite{Park19_DeepSDF}, 
Fourier Feature Networks \cite{tancik2020fourfeat}, SIREN \cite{sitzmann2019siren},
and Neural Implicits (NI) \cite{davies2020overfit}. 
These architectures
show state-of-the-art performance on overfitting 
to 3D shapes and also have source code available. 
We reimplement these baselines 
to the best of our ability using their source code
as references, and provide details in the supplemental material.

\vspace{-1mm}
\paragraph{Mesh Datasets} \tabref{tab:storage} shows overall results across ShapeNet,
Thingi10K, and TurboSquid. 
We sample 150, 32, and 16 shapes respectively
from each dataset, and overfit to each shape using 
100, 100 and 600 epochs respectively. 
For ShapeNet150, we use 50 shapes each from the \textit{car}, 
\textit{airplane} and \textit{chair} categories. For Thingi32, we use 32 shapes tagged as \textit{scans}.
ShapeNet150 and Thingi32 are evaluated using Chamfer-$L^1$ distance 
(multiplied by $10^3$) and intersection over union
over the uniformly sampled points (gIoU). TurboSquid has
much more interesting surface features, so we use
both the 3D geometry-based metrics as well as image-space metrics 
based on 32 multi-view rendered images. 
Specifically, we calculate intersection
over union for the segmentation mask (iIoU) and 
image-space normal alignment 
using $L^2$-distance on the mask intersection. 
The shape complexity roughly increases over the datasets. We 
train 5 LODs for ShapeNet150 and Thingi32, and 6 LODs for TurboSquid. 
For dataset preparation, we follow DualSDF \cite{hao2020dualsdf} and
normalize the mesh, remove internal triangles, and sign the distances with ray stabbing \cite{raystab}.

\input{figures/mesh}

Storage (KB) corresponds to the sum of the decoder size and the 
representation, assuming 32-bit precision. 
For our architecture,
the decoder parameters consist of 90 KB of the storage impact,
so the effective storage size is smaller
for lower LODs since the decoder is able to generalize to multiple shapes. 
The {\it \# Inference Params.} are the 
number of parameters required for the distance query, 
which roughly correlates to the number of flops required for inference.

Across all datasets and metrics, we achieve state-of-the-art results. 
Notably, our representation
shows better results starting at the third LOD, 
where we have minimal storage impact. We also note
our inference costs are fixed at \numprint{4737} floats across all resolutions, 
requiring 99\% less inference parameters
compared to FFN \cite{tancik2020fourfeat} and 37\% less than Neural Implicits \cite{davies2020overfit}, 
while showing better reconstruction quality (see \figref{fig:mesh} for a qualitative evaluation).

\input{figures/old_car}

\paragraph{Special Case Analytic SDFs}
We 
also evaluate reconstruction on 
two particularly difficult analytic SDFs collected from Shadertoy. 
The {\it Oldcar} model is a highly non-metric SDF, which 
does not satisfy the Eikonal equation $|\gradient \sdf| = 1$ and contains discontinuities.
This is a critical case to handle, because 
non-metric SDFs 
are 
often exploited for special effects and easier modeling of SDFs. 
The {\it Mandelbulb} is 
a recursive fractal with infinite resolution. Both SDFs are defined by mathematical
expressions, which we extract and sample distance values from.
We train these analytic shapes for 100 epochs against
$5\times10^6$ samples per epoch.

Only our architecture can capture the high-frequency 
details of these complex
examples to reasonable accuracy. Notably, both 
FFN \cite{tancik2020fourfeat}
and SIREN \cite{sitzmann2019siren} seem to fail entirely; 
this is likely because both
can only fit smooth distance fields and are unable to handle 
discontinuities and recursive structures. 
See \figref{fig:oldcar} for a
qualitative comparison.

\vspace{-1mm}
\paragraph{Convergence} We perform experiments to evaluate training convergence speeds
of our architecture. Table \ref{tab:convergence} shows 
reconstruction results on Thingi32 on our model fully trained for 100 epochs, 
trained for 30 epochs, and trained for 30 epochs from
pretrained weights on the Stanford Lucy statue (\figref{fig:teaser}). We find that our architecture
converges quickly and achieves better reconstruction 
even with roughly 45\% the training time of DeepSDF~\cite{Park19_DeepSDF} and FFN~\cite{tancik2020fourfeat}, which are trained for the full 100 epochs. 
Finetuning from pretrained weights helps with lower LODs, but the difference is small.
Our representation swiftly converges to good solutions.

\input{tables/convergence}

\input{tables/frametimes}

\subsection{Rendering Performance}

We also evaluate the inference performance of our architecture, both with and without 
our rendering algorithm. We first evaluate the performance using a naive Python-based sphere tracing algorithm 
in PyTorch \cite{paszke2019pytorch}, with the same 
implementation across all baselines for fair comparison. 
For the Python version of our representation, we store the features on a dense voxel grid, 
since a naive sphere tracer cannot handle sparsity. 
For the optimized implementation, we show the
performance of our representation using a renderer implemented using 
libtorch \cite{paszke2019pytorch}, CUB \cite{merrill2017cub}, and CUDA.

\tabref{tab:frametime} shows frametimes on the TurboSquid {\it V Mech} scene with a 
variety of different resolutions. Here, we measure frametime as the CUDA time 
for the sphere trace and normal computation. The {\it \# Visible Pixels} 
column shows the number of pixels occupied in the image by the model.
We see that both our naive PyTorch renderer and sparse-optimized CUDA renderer
perform better than the baselines. In particular,
the sparse frametimes are more than 100$\times$ faster than DeepSDF while achieving
better visual quality with less parameters. We also notice
that our frametimes decrease significantly as LOD decreases for our naive renderer 
but less so for our optimized renderer. This is because
the bottleneck of the rendering is not in the ray-octree intersect---which is 
dependent on the number of voxels---but rather in the MLP inference and miscellaneous
memory I/O. We believe there is still significant room for improvement by caching 
the small MLP decoder to minimize data movement. 
Nonetheless, the lower LODs still benefit from lower memory consumption and storage. 

\input{tables/generalization}

\subsection{Generalization}

We now show that our surface extraction mechanism can generalize to multiple shapes,
even from being trained on a {\it single} shape. This is important because loading
distinct weights per object as in~\cite{davies2020overfit, sitzmann2019siren}
incurs large amounts of memory movement, which is expensive. 
With a general surface extraction mechanism, the weights can be pre-loaded
and multi-resolution voxels can be streamed-in on demand.

\tabref{tab:generalization} shows results on Thingi32. 
DeepSDF \cite{Park19_DeepSDF}, FFN \cite{tancik2020fourfeat} and Ours (overfit)
are all overfit per shape. Ours (general) first overfits the architecture 
on the Stanford Lucy model, fixes the surface extraction network weights, and
trains only the sparse features. We see that our 
representation fares better, even against large networks
that are overfitting to each specific shape examples. At the lowest LOD, the surface extractor
struggles to reconstruct good surfaces, as expected; the features become increasingly
high-level and complex for lower LODs.

\input{figures/edge_collapse}

\subsection{Geometry Simplification}

In this last experiment, we evaluate how our low LODs perform against  
classic mesh decimation algorithms, in particular edge collapse \cite{garland_heckbert}
in libigl \cite{libigl}. 
We compare against mesh decimation instead of mesh compression algorithms \cite{alliez2005recent}
because our model can also benefit from compression and {\it mesh decoding} incurs
additional runtime cost.
We first evaluate our memory impact,
which $M = \,(\featureDim+1)\,|\allVoxels_{1:\maxLOD}|$ bytes where $\featureDim+1$ is the feature
dimension along with the {$Z$-order} curve \cite{morton1966computer} for indexing, 
and $|\allVoxels_{1:\maxLOD}|$ is the octree size. 
We then calculate the face budget as $\faces = M / 3$ since the connectivity
can be arbitrary. As such, we choose a conservative budget to benefit the mesh representation. 

\figref{fig:edgecollapse} shows results on the Lion statue from Thingi32.
We see that as the memory budget decreases, the relative advantage
on the perceptual quality of our method increases, as evidenced by
the image-based normal error.
The SDF can represent 
smooth features easily, whereas the mesh suffers from discretization errors 
as the face budget decreases. Our representation can also smoothly 
blend between LODs by construction, something difficult to do with meshes.

%% file: tables/storage.tex
\bgroup
\def\arraystretch{1.0}
\begin{table*}[t]
\begin{center}
\resizebox{\linewidth}{!}{%
\rowcolors{2}{gray!15}{white}
\begin{tabular}{lcccccccccccc}
\toprule
\rowcolor{white}
&&& \multicolumn{2}{c}{ShapeNet150 \cite{chang2015shapenet}} && \multicolumn{2}{c}{Thingi32 \cite{zhou2016thingi10k}} && \multicolumn{4}{c}{TurboSquid16}  \\
\cmidrule{4-5} \cmidrule{7-8} \cmidrule{10-13} 
\rowcolor{white}
& Storage (KB) & \# Inference Param. & gIoU $\uparrow$ & Chamfer-$L^1$ $\downarrow$ && gIoU $\uparrow$ & Chamfer-$L^1$ $\downarrow$ && iIoU $\downarrow$ & Normal-$L^2$ $\downarrow$ & gIoU $\uparrow$ & Chamfer-$L^1$ $\downarrow$\\
\midrule
DeepSDF \cite{Park19_DeepSDF}             & \numprint{7186} & \numprint{1839614} & 86.9 & 0.316 && 96.8 & 0.0533 && 97.6 & 0.180 & 93.7 & 0.211  \\
FFN \cite{tancik2020fourfeat}             & \numprint{2059} & \numprint{526977}  & 88.5 & 0.077   && 97.7 & 0.0329    && 95.5 & 0.181 & 92.2 & 0.362 \\
SIREN \cite{sitzmann2019siren}            & \numprint{1033} & \numprint{264449}  & 78.4 & 0.381 && 95.1 & 0.0773 && 92.9 & 0.208 & 82.1 & 0.488 \\
Neural Implicits \cite{davies2020overfit} & {\bf30}   & \numprint{7553}          & 82.2 & 0.500 && 96.0 & 0.0919 && 93.5 & 0.211 & 82.7 & 0.354  \\
\midrule
Ours / LOD 1 & 96   & \numprint{4737}                                            & 84.6 & 0.343  && 96.8 & 0.0786       && 91.9	&	0.224	& 	79.7	&	0.471 \\
Ours / LOD 2 & 111  & \numprint{4737}                                            & 88.3 & 0.198  && 98.2 & 0.0408       && 94.2	&	0.201	& 	87.3	&	0.283 \\
Ours / LOD 3 & 163  & \numprint{4737}                                            & 90.4 & 0.112  && 99.0 & 0.0299       && 96.1	&	0.184	& 	91.3	&	0.162 \\
Ours / LOD 4 & 391  & \numprint{4737}                                            & 91.6 & 0.069  && 99.3 & 0.0273       && 97.1	&	0.170	& 	94.3	&	0.111 \\
Ours / LOD 5 & \numprint{1356} & \numprint{4737}                                 & {\bf 91.7} & {\bf 0.062 }&& {\bf 99.4}  & {\bf 0.0271} && 98.3	&	0.166	& 	95.8	&	0.085 \\
Ours / LOD 6 & \numprint{9826} & \numprint{4737}                                 & -- & -- && -- & --                 && \textbf{98.5}	&	\textbf{0.167}	& 	\textbf{96.7}	&	\textbf{0.076} \\
\bottomrule
\end{tabular}
}
\end{center}
\vspace{-1.2em}
\caption{\textbf{Mesh Reconstruction.} This table shows architectural 
and per-shape reconstruction comparisons against three different datasets.  
We see that under all evaluation schemes, our architecture starting from LOD 3 performs
much better despite having much lower storage and inference parameters. 
The storage for our representation
is calculated based on the average sparse voxel counts across all shapes in all datasets plus the decoder size,
and \textit{\# Inference Param.} measures network parameters used for a single distance query.}
\label{tab:storage}
\vspace{-2mm}
\end{table*}
\egroup

%% file: figures/mesh.tex
\begin{figure}[t]
    \centering
    \vspace{-5pt}
    \begin{subfigure}[t]{0.24\linewidth}
        \centering
        \includegraphics[width=\textwidth]{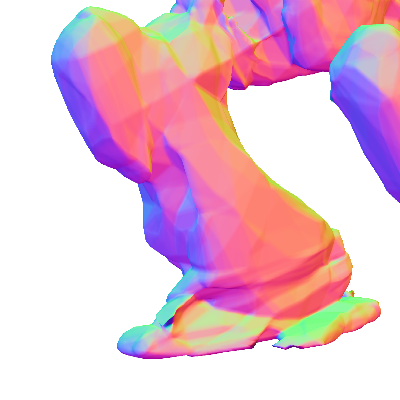}
        \includegraphics[width=\textwidth]{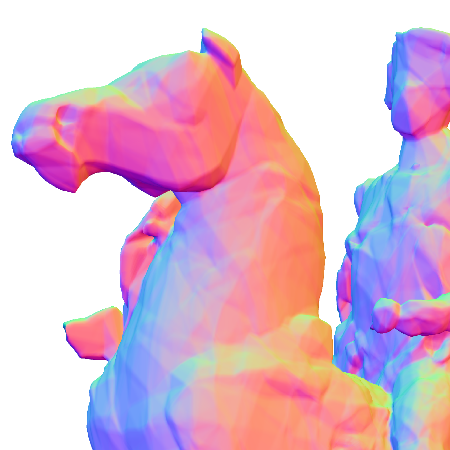}
        \includegraphics[width=\textwidth]{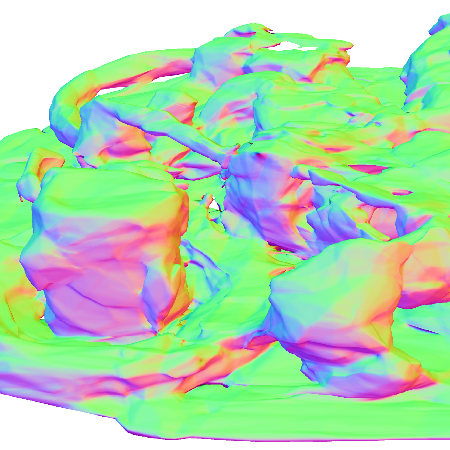}
        \caption*{NI \cite{davies2020overfit}}
    \end{subfigure}
    \begin{subfigure}[t]{0.24\linewidth}
        \centering
        \includegraphics[width=\textwidth]{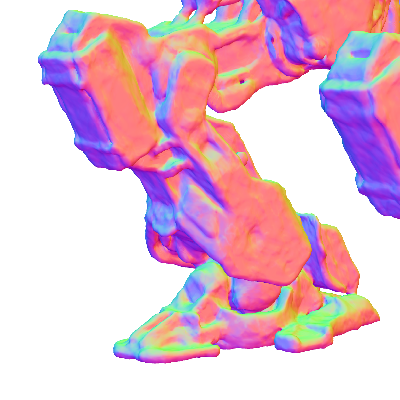}
        \includegraphics[width=\textwidth]{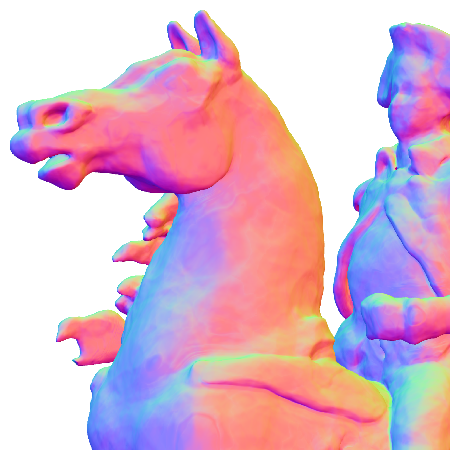}
        \includegraphics[width=\textwidth]{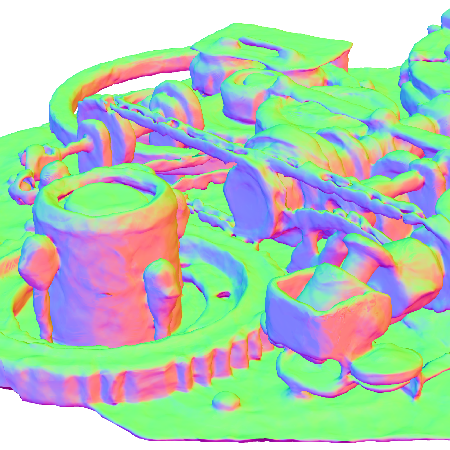}
        \caption*{FFN \cite{tancik2020fourfeat}}
    \end{subfigure}
    \begin{subfigure}[t]{0.24\linewidth}
        \centering
        \includegraphics[width=\textwidth]{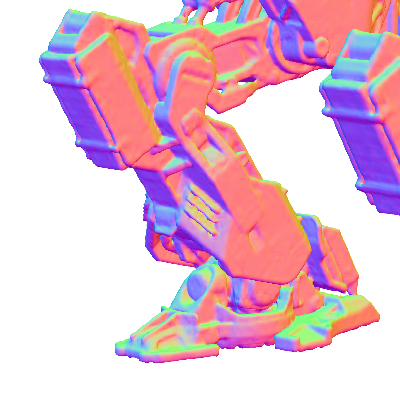}
         \includegraphics[width=\textwidth]{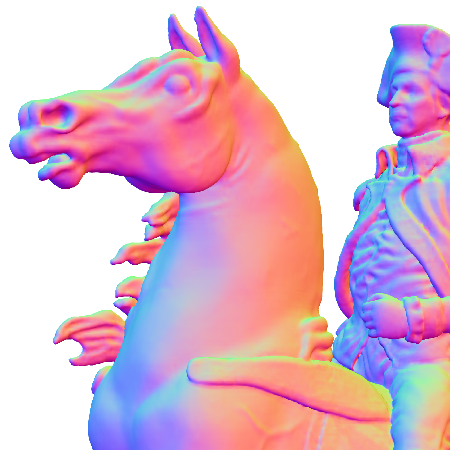}
         \includegraphics[width=\textwidth]{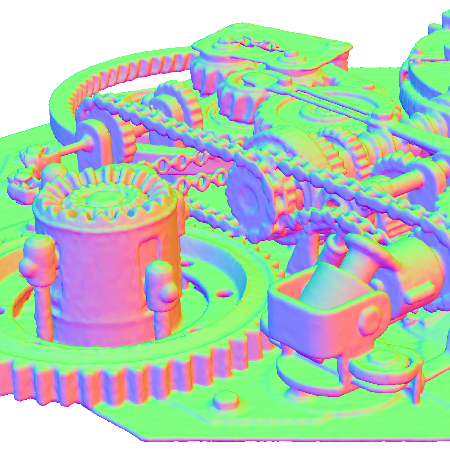}
        \caption*{Ours / LOD 6}
    \end{subfigure}
    \begin{subfigure}[t]{0.24\linewidth}
        \centering
        \includegraphics[width=\textwidth]{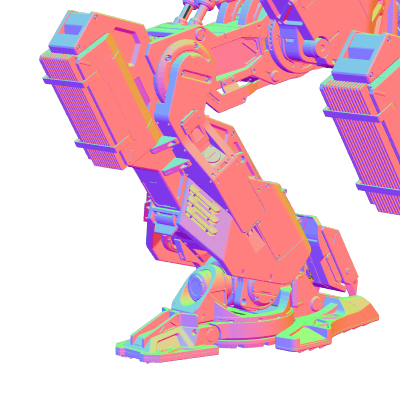}
         \includegraphics[width=\textwidth]{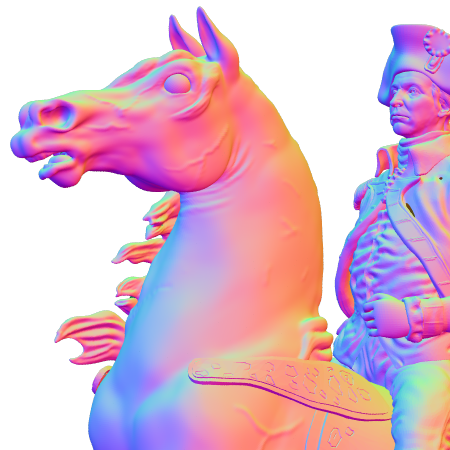}
         \includegraphics[width=\textwidth]{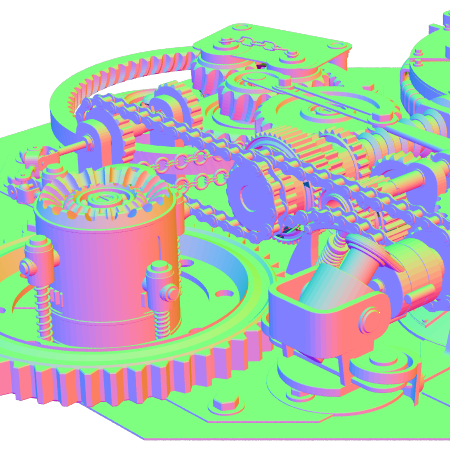}
        \caption*{Reference}
    \end{subfigure}
    \vspace{-2mm}
    \caption{\textbf{Comparison on TurboSquid.} We qualitatively compare the mesh reconstructions. Only ours 
    is able to recover fine details, with speeds 50$\times$ faster than FFN and comparable to NI.
    We render surface normals
    to highlight geometric details.}
\label{fig:mesh}
\vspace{-1.25em}
\end{figure}

%% file: figures/old_car.tex
\begin{figure*}[t]
    \centering
    \vspace{-8pt}
    \begin{subfigure}[t]{0.16\linewidth}
        \centering
        \includegraphics[width=\textwidth]{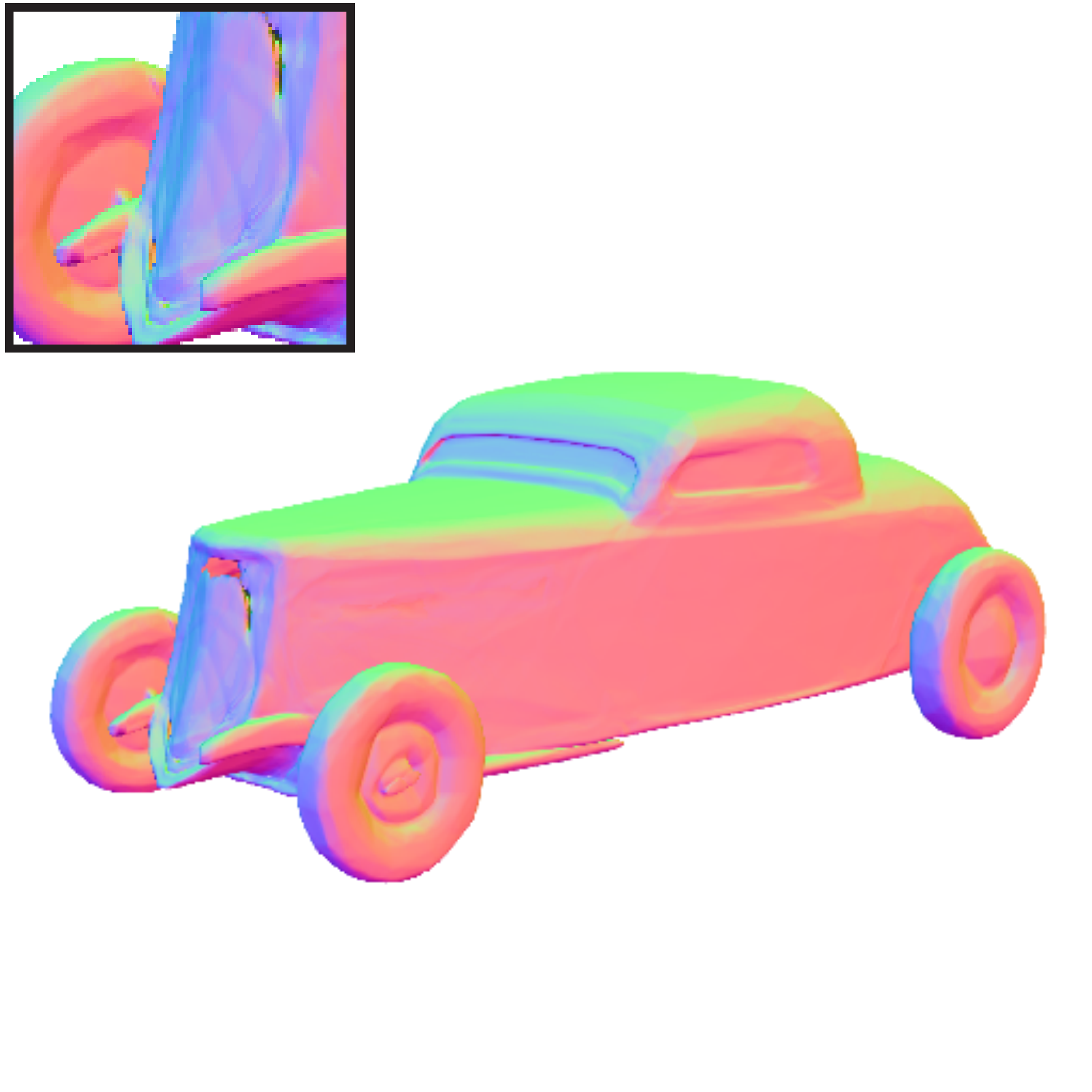}\vspace{-10pt}
        \includegraphics[width=\textwidth]{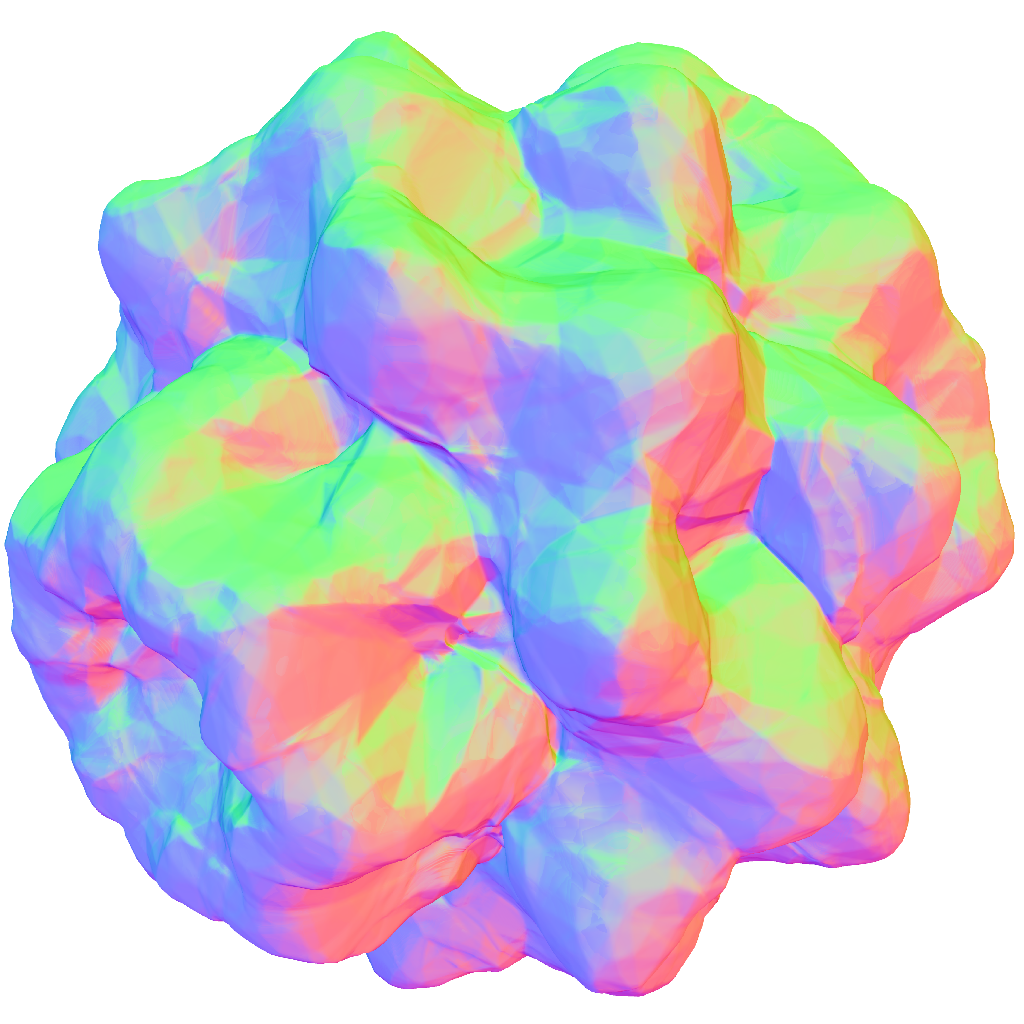}
        \caption*{DeepSDF \cite{Park19_DeepSDF}}
    \end{subfigure}%
    \begin{subfigure}[t]{0.16\linewidth}
        \centering
        \includegraphics[width=\textwidth]{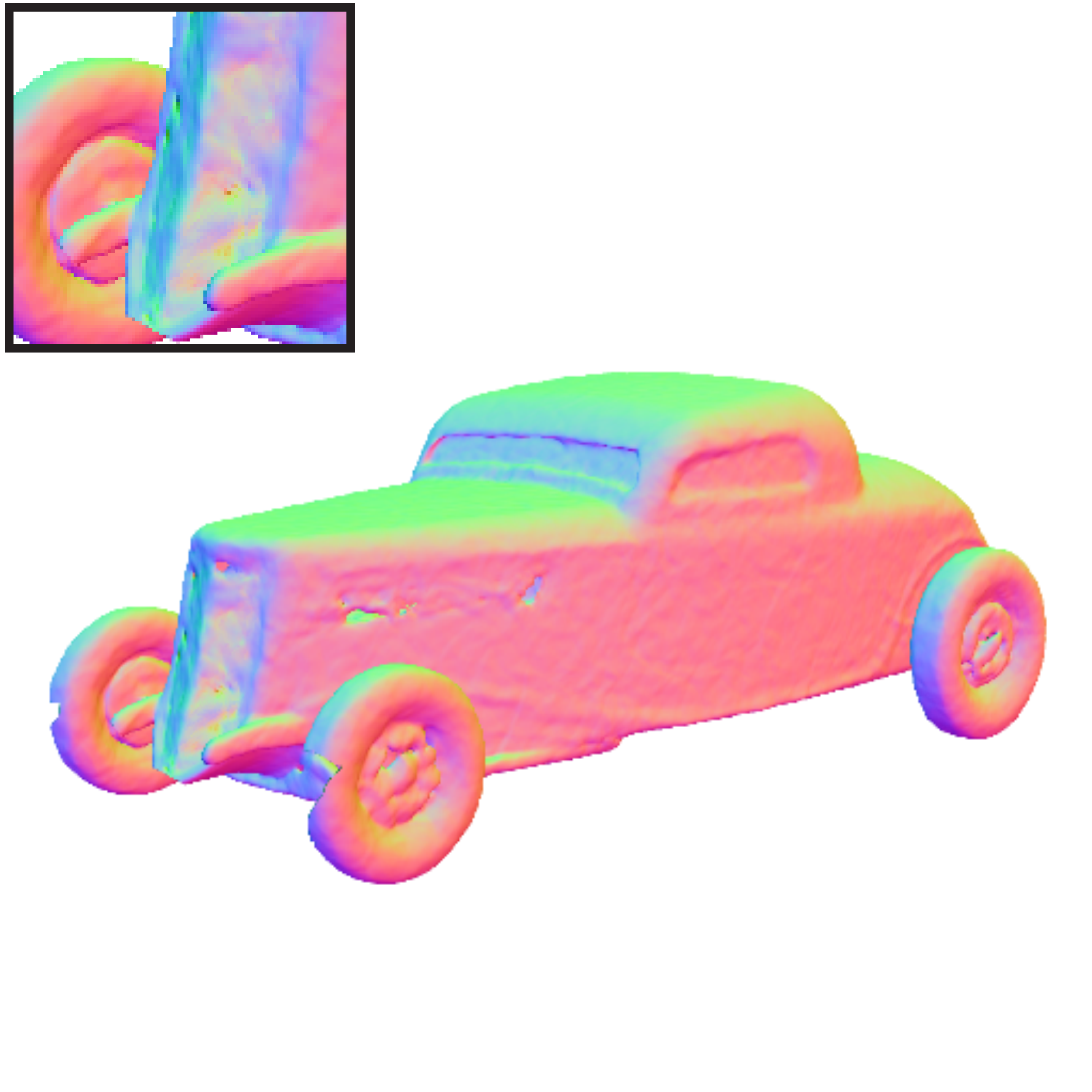}\vspace{-10pt}
        \includegraphics[width=\textwidth]{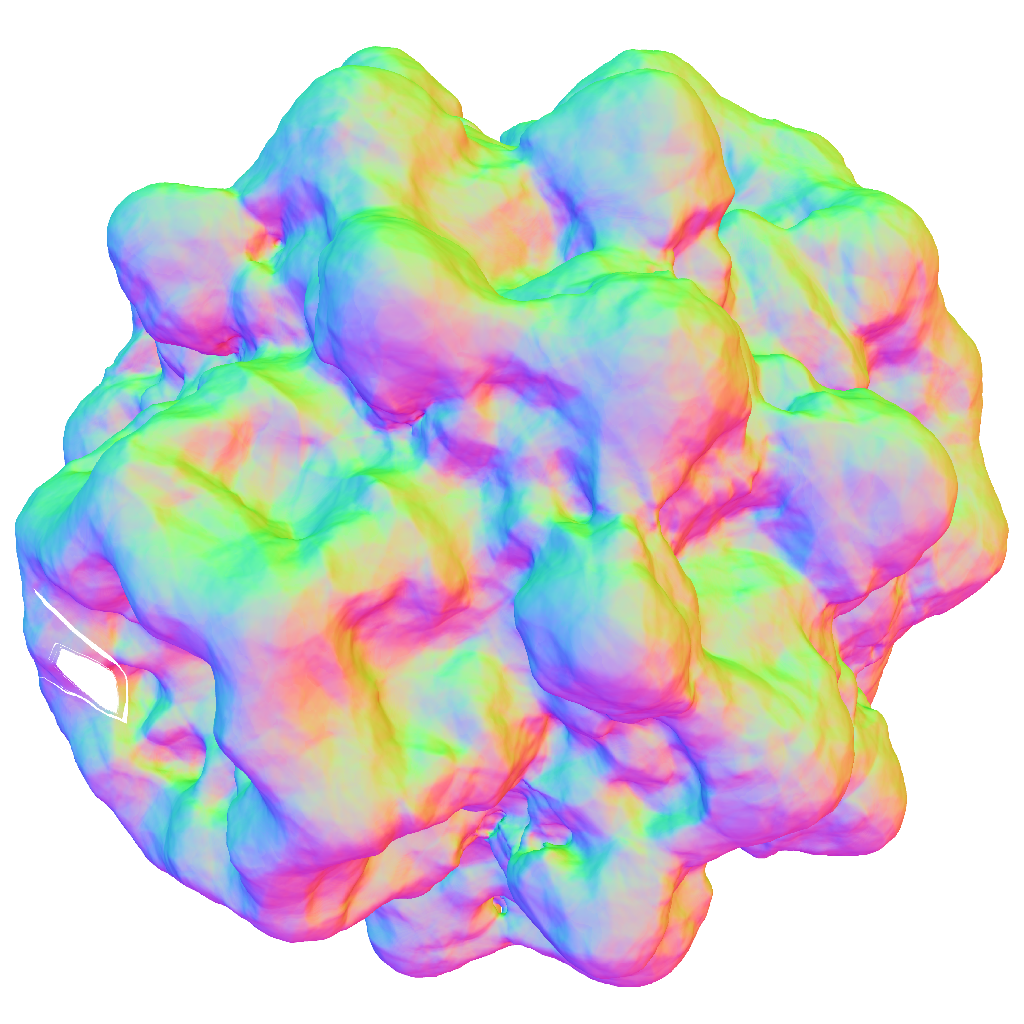}
        \caption*{FFN \cite{tancik2020fourfeat}}
    \end{subfigure}%
    \begin{subfigure}[t]{0.16\linewidth}
        \centering
        \includegraphics[width=\textwidth]{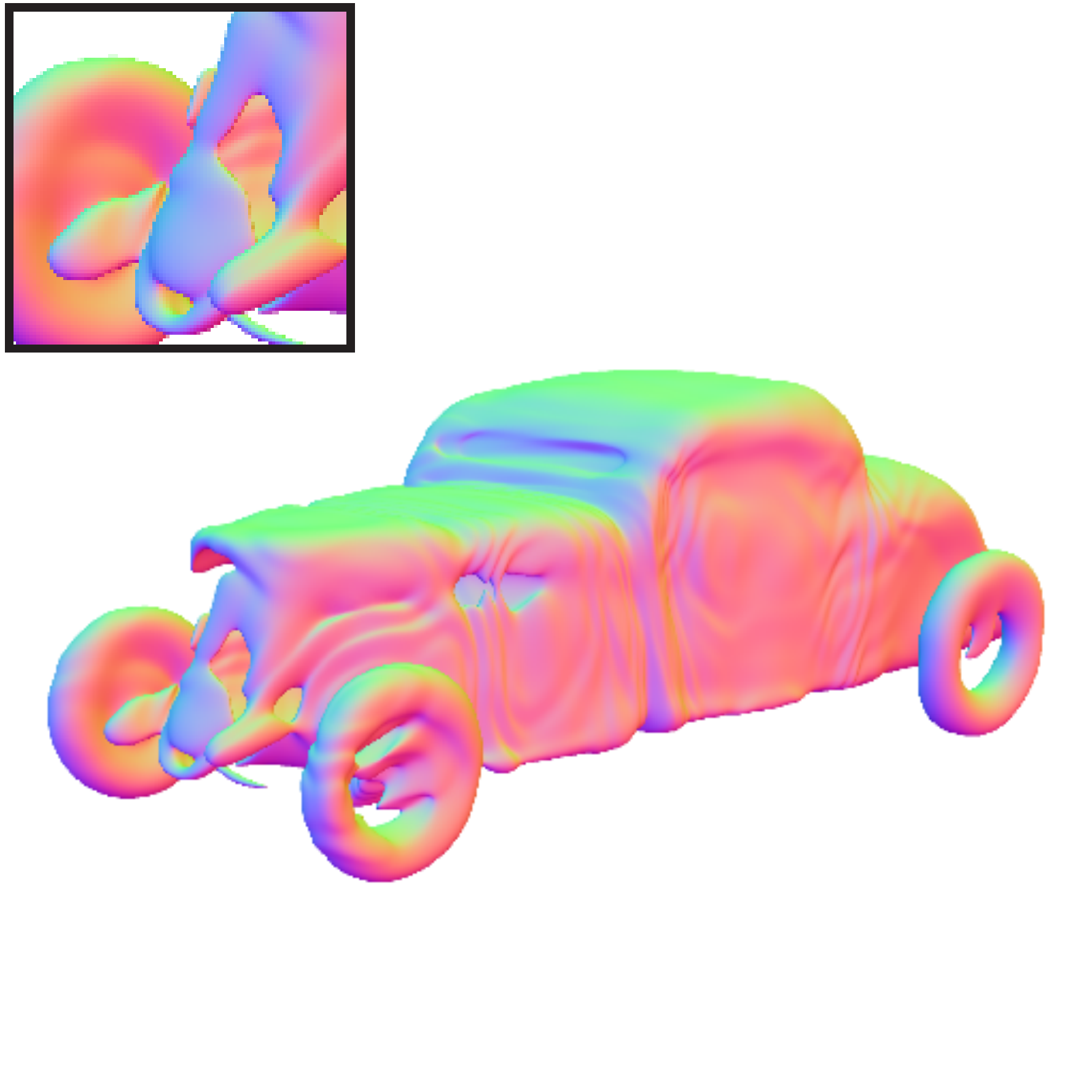}\vspace{-10pt}
        \includegraphics[width=\textwidth]{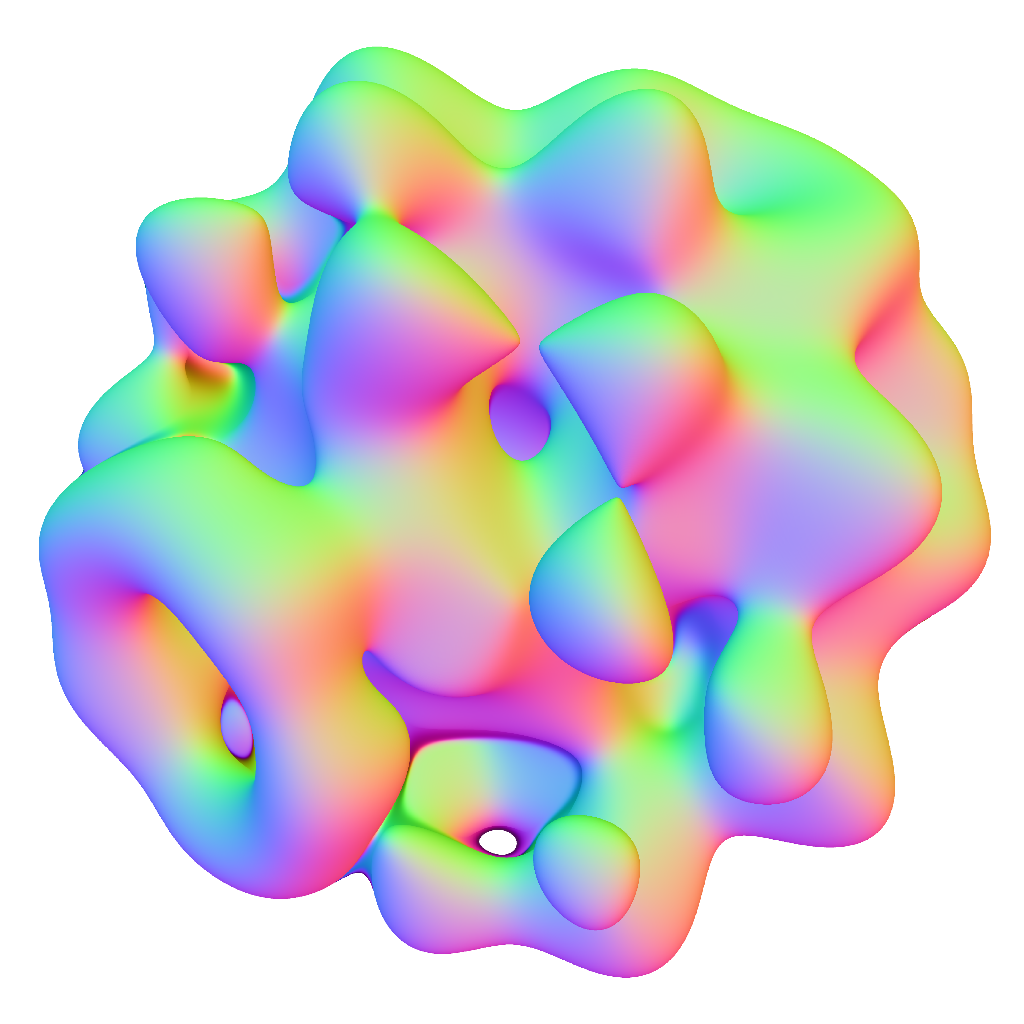}
        \caption*{SIREN \cite{sitzmann2019siren}}
    \end{subfigure}%
    \begin{subfigure}[t]{0.16\linewidth}
        \centering
        \includegraphics[width=\textwidth]{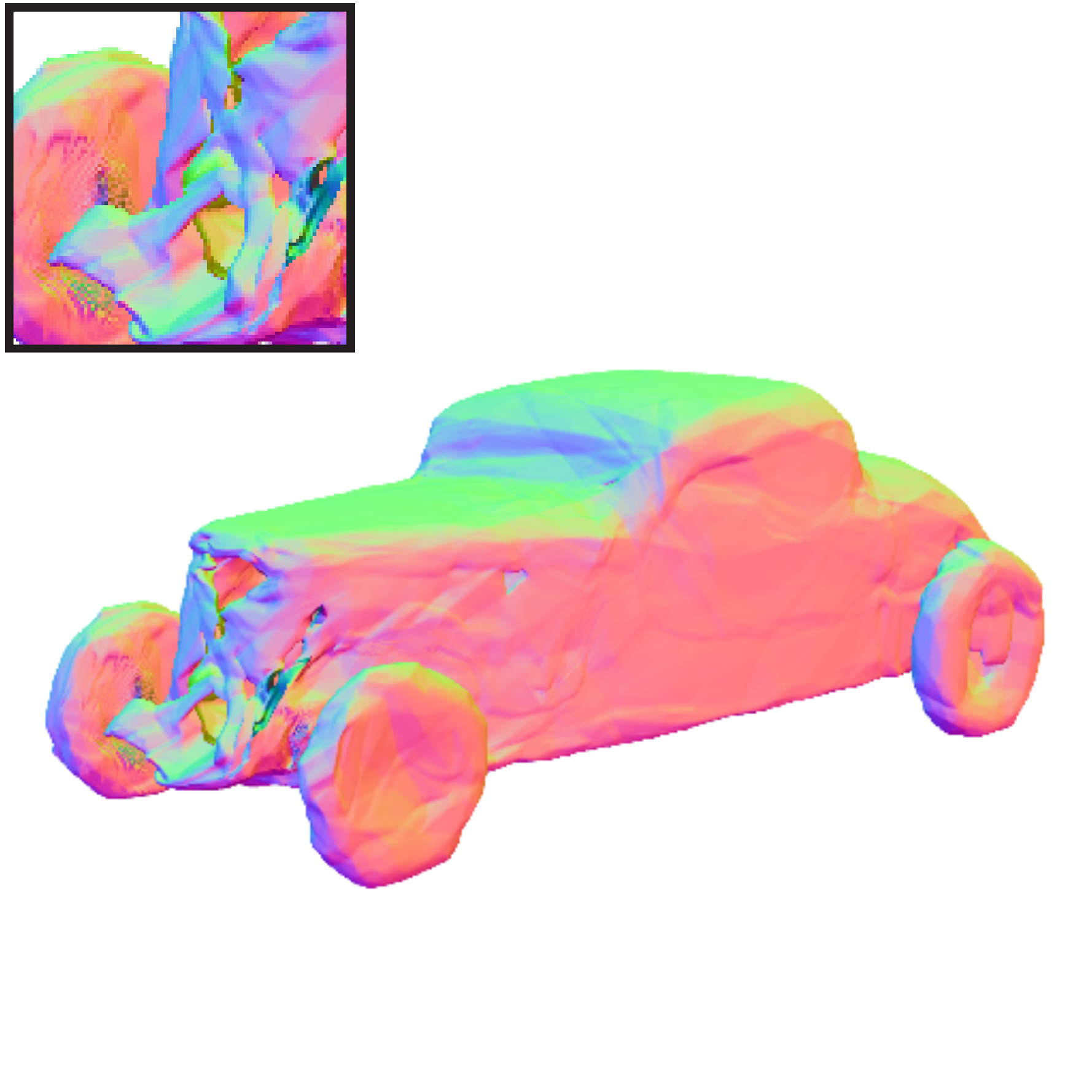}\vspace{-10pt}
        \includegraphics[width=\textwidth]{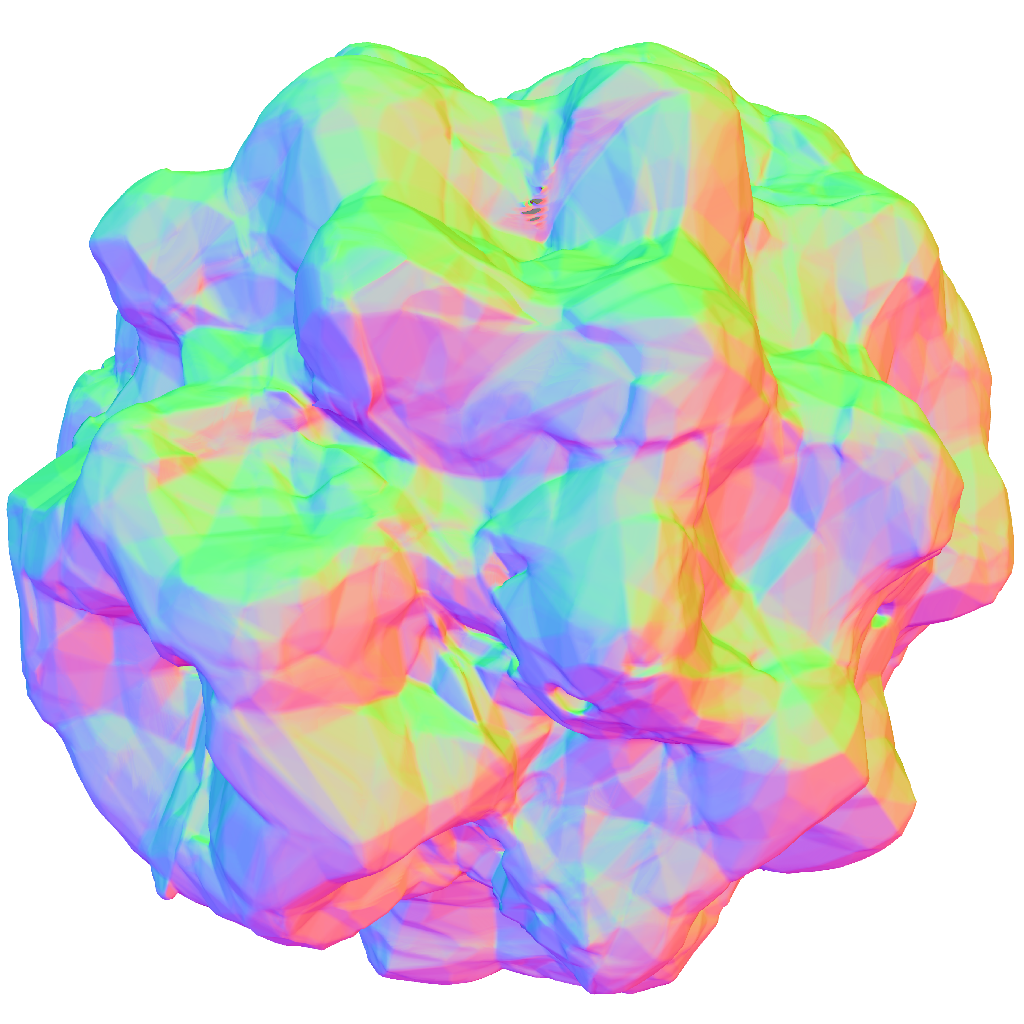}
        \caption*{Neural Implicits \cite{davies2020overfit}}
    \end{subfigure}%
    \begin{subfigure}[t]{0.16\linewidth}
        \centering
        \includegraphics[width=\textwidth]{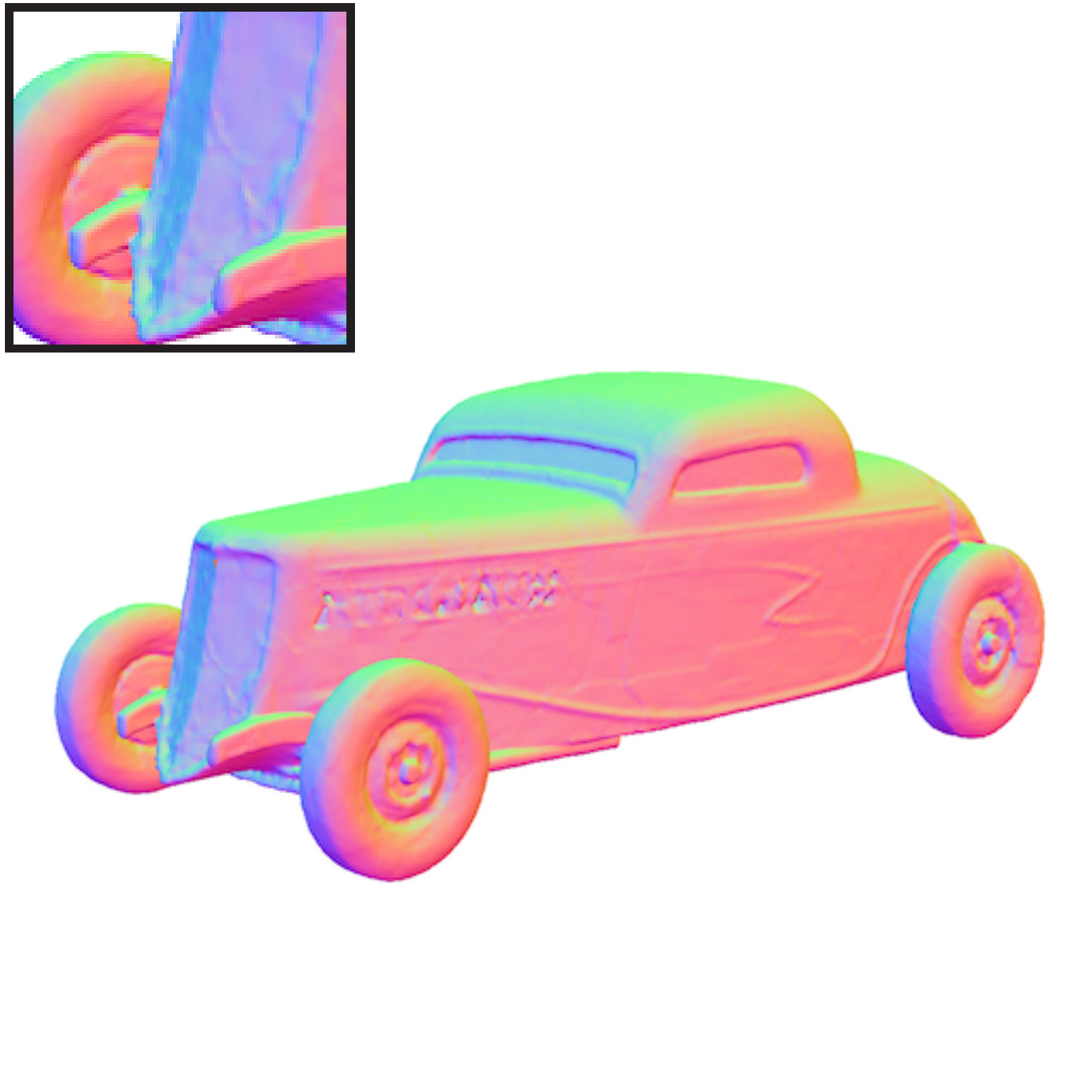}\vspace{-10pt}
        \includegraphics[width=\textwidth]{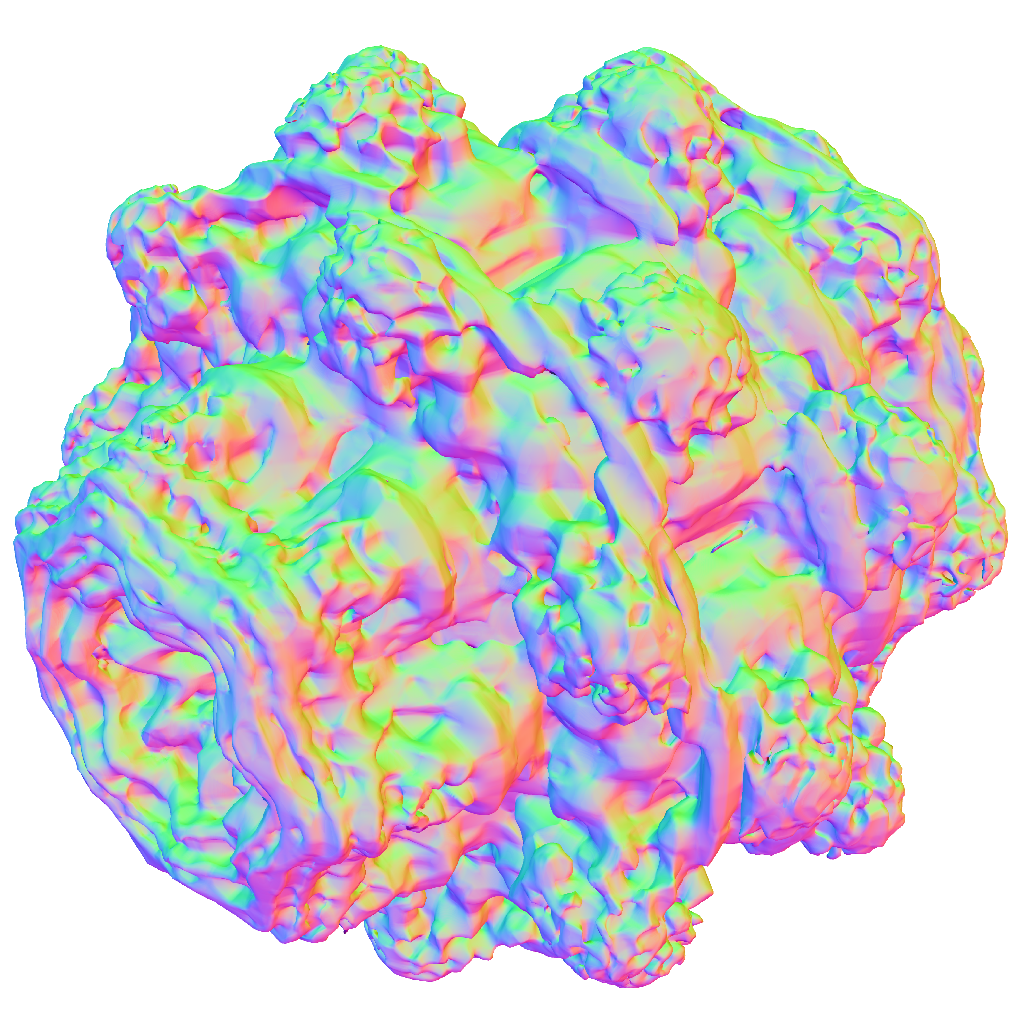}
        \caption*{Ours / LOD 5}
    \end{subfigure}
    \begin{subfigure}[t]{0.16\linewidth}
        \centering
        \includegraphics[width=\textwidth]{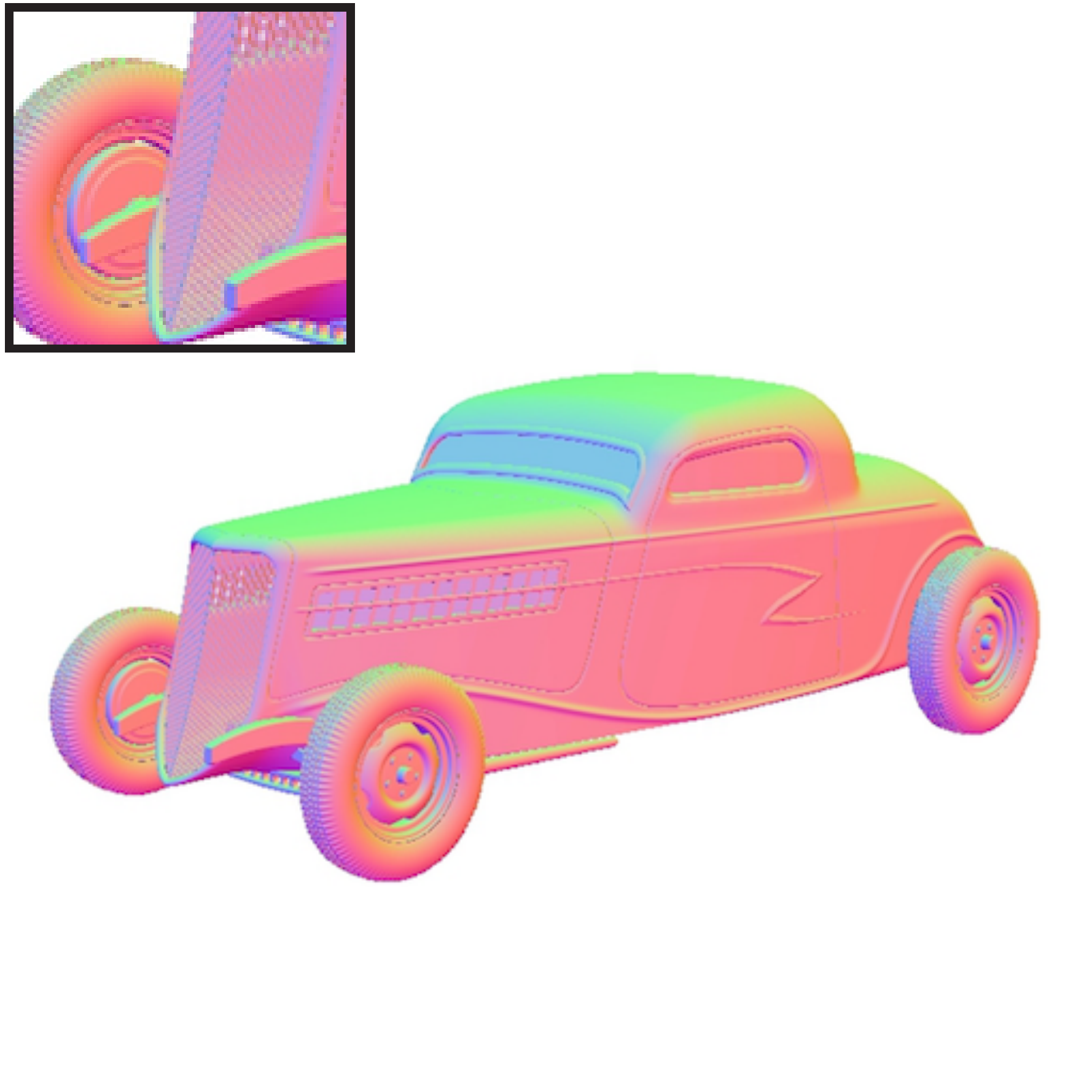}\vspace{-10pt}
        \includegraphics[width=\textwidth]{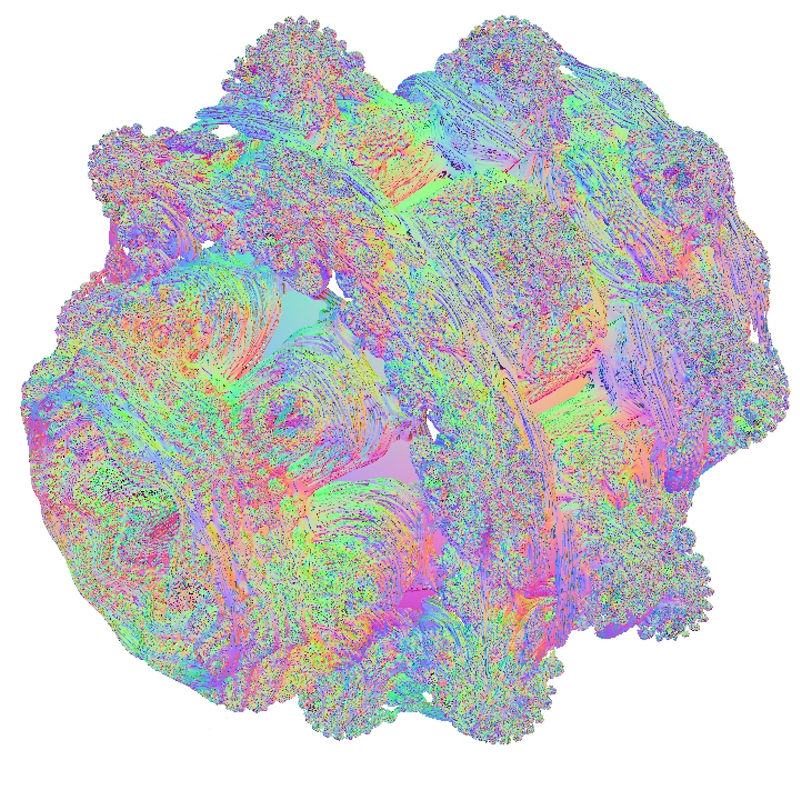}
        \caption*{Reference}
    \end{subfigure}
    \vspace{-1.8mm}
    \caption{\textbf{Analytic SDFs.} We test against two difficult analytic SDF
    examples from Shadertoy; the \textit{Oldcar}, which contains a highly non-metric signed distance field,
    as well as the \textit{Mandelbulb}, which is a recursive fractal structure that can only be expressed using implicit surfaces. 
    Only our architecture can reasonably reconstruct these hard cases. We render surface normals
    to highlight geometric details.}
\label{fig:oldcar}
\vspace{-9pt}
\end{figure*}

%% file: tables/convergence.tex
\bgroup
\def\arraystretch{1.0}
\begin{table}[t]
\begin{center}
\resizebox{\linewidth}{!}{%
\rowcolors{2}{gray!15}{white}
\begin{tabular}{lcccccccc}
\toprule
\rowcolor{white}
Method \, / \, LOD & 1 & 2 & 3 & 4 & 5  \\
\midrule
DeepSDF (100 epochs) \cite{Park19_DeepSDF} & 0.0533 & 0.0533 & 0.0533 & 0.0533 & 0.0533 \\
FFN (100 epochs) \cite{tancik2020fourfeat} & \textbf{0.0329} & \textbf{0.0329} & 0.0329 & 0.0329 & 0.0329   \\
\midrule
Ours (30 epochs) & 0.1197 & 0.0572 & 0.0345 & 0.0285 & 0.0278 \\
Ours (30 epochs, pretrained)  & 0.1018 & 0.0499 & 0.0332 & 0.0287 & 0.0279 \\
Ours (100 epochs)      & 0.0786 & 0.0408 & \textbf{0.0299} & \textbf{0.0273} & \textbf{0.0271} \\
\bottomrule
\end{tabular}
}
\end{center}
\vspace{-5mm}
\caption{\textbf{Chamfer-$L^1$ Convergence.} 
We evaluate the performance of our architecture on
the Thingi32 dataset under
different training settings and report faster convergence for higher LODs.} 
\label{tab:convergence}
\vspace{-4mm}
\end{table}
\egroup

%% file: tables/frametimes.tex
\bgroup
\def\arraystretch{1.0}
\begin{table*}[t]
\begin{center}
\resizebox{\linewidth}{!}{%
\rowcolors{2}{gray!15}{white}
\begin{tabular}{ccccccccccc}
\toprule
& & \multicolumn{8}{c}{Frametimes (ms) \,/\, Improvement Factor} \\
\cmidrule{3-10}
\rowcolor{white}
Resolution & \# Visible Pixels & DeepSDF \cite{Park19_DeepSDF} & FFN \cite{tancik2020fourfeat} & SIREN \cite{sitzmann2019siren} & NI \cite{davies2020overfit} & Ours (N) / LOD 4 & Ours (N) / LOD 6 & Ours (S) / LOD 4  & Ours (S) / LOD 6   \\
\midrule
640 $\times$ 480 & \numprint{94624}  & \numprint{1693} \,/\, 57$\times$ & \numprint{1058} \,/\, 35$\times$ & 595 \,/\, 20$\times$ & 342 \,/\, 11$\times$ & 164 \,/\, 5$\times$ & 315 \,/\, 11$\times$ & 28 & 30 \,/\, 1$\times$  \\
\numprint{1280} $\times$ \numprint{720} & \numprint{213937} & \numprint{4901} \,/\, 96$\times$ & \numprint{2760} \,/\, 54$\times$ & \numprint{1335} \,/\, 26 $\times$ & 407 \,/\, 8$\times$ & 263 \,/\, 5$\times$ & 459 \,/\, 9$\times$ & 50 & 51 \,/\, 1$\times$ \\
\numprint{1920} $\times$ \numprint{1080} & \numprint{481828} &  \numprint{10843} \,/\, 119$\times$ & \numprint{5702} \,/\, 62$\times$ & \numprint{2946} \,/\, 32$\times$ & 701 \,/\, 8$\times$  & 473 \,/\, 5$\times$  & 784 \,/\, 9$\times$ & 93 & 91 \,/\, 1$\times$ \\

\bottomrule
\end{tabular}
}
\end{center}
\vspace{-5mm}
  \caption{\textbf{Rendering Frametimes.} 
  We show runtime comparisons between different representations, where (N) and (S) correspond to our naive and sparse renderers, respectively.
  We compare baselines against Ours (Sparse) at LOD 6.
  \# Visible Pixels shows the number of 
  pixels occupied by the benchmarked scene (TurboSquid \textit{V Mech}), 
  and frametime measures ray-tracing and surface normal computation.}
\label{tab:frametime}
\vspace{-3mm}
\end{table*}

%% file: tables/generalization.tex
\bgroup
\def\arraystretch{1.0}
\begin{table}[t]
\begin{center}
\resizebox{\linewidth}{!}{%
\rowcolors{2}{gray!15}{white}
\begin{tabular}{lcccccccc}
\toprule
\rowcolor{white}
& \multicolumn{5}{c}{Chamfer-$L^1$ $\downarrow$}\\ 
\cmidrule{2-6}
\rowcolor{white}
Method \, / \, $\lodd$ & 1 & 2 & 3 & 4 & 5  \\
\midrule
DeepSDF (overfit per shape)  \cite{Park19_DeepSDF}& 0.0533 & 0.0533 & 0.0533 & 0.0533 & 0.0533 \\
FFN (overfit per shape) \cite{tancik2020fourfeat} & 0.0322 & 0.0322 & 0.0322 & 0.0322 & 0.0322 \\
\midrule
Ours (overfit per shape) &0.0786 & 0.0408 & 0.0299 & 0.0273 & 0.0271 \\
Ours (general)   &0.0613 & 0.0378 & 0.0297 & 0.0274 & 0.0272 \\
\bottomrule
\end{tabular}
}
\end{center}
\vspace{-5mm}
\caption{\textbf{Generalization.} 
We evaluate generalization on
Thingi32. Ours (general) freezes surface extractor weights
pretrained on a {\it single} shape, 
and only trains the feature volume. Even against large overfit networks, 
we perform better at high LODs.} 
\label{tab:generalization}
\vspace{-2pt}
\end{table}
\egroup

%% file: figures/edge_collapse.tex
\begin{figure}[t]
    \centering
    \includegraphics[width=1.\linewidth]{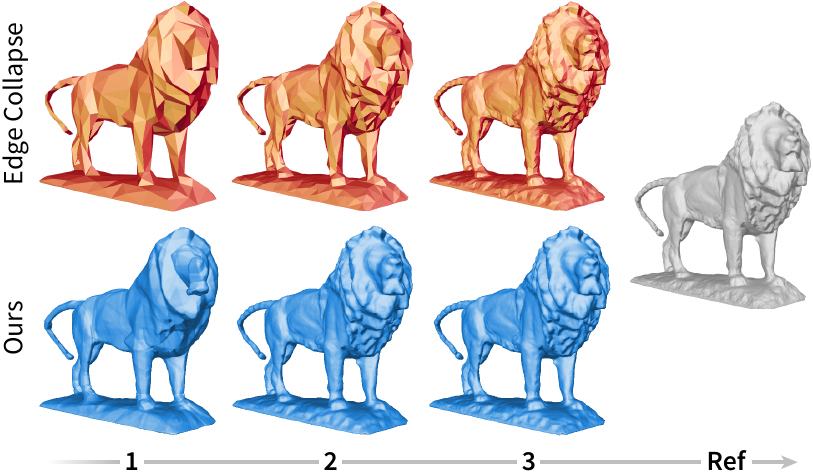}
    \vspace{-8pt}
    
    \resizebox{\linewidth}{!}{%
    \begin{tabular}{lccccc}
    \toprule
    & $\lodd$ & gIoU $\uparrow$ & Chamfer-$L^1$ $\downarrow$ & iIoU $\uparrow$ & Normal-$L^2$ $\downarrow$ \\
    \midrule
    & 1 & 94.4 & 0.052 & 97.4 & 0.096\\
    \rowcolor{gray!15}
    Decimation \cite{garland_heckbert} & 2 & 97.4 & 0.026 & 98.7 & 0.069\\
    & 3 & 99.1 & 0.019 & 99.5 & 0.044 \\
    \midrule
    & 1 & 96.0 / $+$1.6 & 0.063 / $+$0.011 & 96.4 / $-$1.0  & 0.044 / $-$0.052 \\
    \rowcolor{gray!15}
    Ours & 2 & 97.8 / $+$0.4  & 0.030 / $+$0.004  & 97.6 / $-$1.1  & 0.035 / $-$0.034 \\
    & 3 & 98.8 / $-$0.3  & 0.023 / $+$0.004 & 98.2 / $-$1.3  & 0.030 / $-$0.014 \\
    \bottomrule
    \end{tabular}
    }
    \vspace{-2mm}
    \caption{\textbf{Comparison with Mesh Decimation.} At low memory budgets, our model
    is able to maintain visual details better than mesh decimation, as seen from lower normal-$L^2$ error.}
\label{fig:edgecollapse}
\vspace{-4mm}
\end{figure}

%% file: sections/4_conclusion.tex
\section{Limitations and Future Work} 

In conclusion, we introduced \textit{Neural Geometric LOD}, a representation for implicit 3D shapes
that achieves state-of-the-art geometry reconstruction quality while allowing real-time rendering with acceptable memory footprint. Our model combines a small surface extraction neural network with a sparse-octree data structure that encodes the geometry and naturally enables LOD. Together with a tailored sphere tracing algorithm, this results in a method that is both computationally performant and highly expressive.

Our approach heavily depends on the point samples used during training. Therefore, scaling our representation to extremely large scenes or very thin, volume-less geometry is difficult. Furthermore, we are not able to easily animate or deform our geometry using traditional methods. We identify these challenges as promising directions for future research. Nonetheless, we believe our model represents a major step forward in neural implicit function-based geometry, being, to the best of our knowledge, the first representation of this kind that can be rendered and queried in real-time. We hope that it will serve as an important component for many downstream applications, such as
scene reconstruction, ultra-precise robotics path planning, interactive content creation, and more.

\paragraph{Acknowledgements} 
We thank Jean-Francois Lafleche, Peter Shirley, Kevin Xie, Jonathan Granskog, Alex Evans, 
and Alex Bie for interesting discussions throughout the project. We also thank Jacob Munkberg, Peter Shirley, Alexander Majercik, David Luebke, Jonah Philion, and Jun Gao for help with paper review.

%% file: sections/5_supp.tex

\appendix
\section{Implementation Details}

\subsection{Architecture}
We set the hidden dimension for all (single hidden layer) MLPs to  
$\hiddenDim=128$. We use a ReLU 
activation function for the intermediate layer and none for the output layer, to support arbitrary distances. 
We set the feature dimension for the SVO to $\featureDim=32$ and initialize all voxel features $\shape \in\shapeSpace$ using a Gaussian prior with $\sigma=0.01$.
We performed ablations and discovered that we get satisfying 
quality with feature dimensions as low as $\featureDim=8$, but we keep $\featureDim=32$ 
as we make bigger gains in storage efficiency
by keeping the octree depth shallower than we save by reducing the feature dimension.

The resolution of each level of the SVO is defined as $\voxelRes_\lodd \eqdef \initRes\cdot 2^\lodd$,
where $\initRes=4$ is the initial resolution, capped at $\maxLOD\in\{5,6\}$ depending
on the complexity of the geometry.
Note that the octree used for rendering  (compare Section \ref{subsec:rendering}) starts at an initial resolution of $1^3$, but we
do not store any feature vectors until the octree reaches a level where the resolution $\initRes=4$.
Each level contains a maximum of ${\voxelRes_\lodd}^3$ voxels.
In practice, the total number is much lower because surfaces are sparse in $\mathbb{R}^3$,
and we only allocate nodes where there is a surface. 

\subsection{Sampling}

We implement a variety of sampling schemes for the generation of our pointcloud datasets.

\paragraph{Uniform} We first sample uniform random positions 
in the bounding volume 
$\boundingVol = [-1,1]^3$ by sampling three uniformly distributed random numbers.

\paragraph{Surface} We have two separate sampling algorithms,
one for meshes and one for signed distance functions.
For meshes, we first compute per-triangle areas.
We then select random triangles with a distribution proportional to the triangle areas,
and then select a random point on the triangle using three uniformly distributed
random numbers and barycentric coordinates. 
For signed distance functions, we first sample uniformly distributed
points in $\boundingVol$. 
We then choose random points on a sphere to form a ray,
and test if the ray hits the surface with sphere tracing.
We continue sampling rays until we find enough rays that hit the surface.

\paragraph{Near} We can additionally sample near-surface points of a mesh
by taking the surface samples, and perturbing the vector with
random Gaussian noise with $\sigma=0.01$. 

\subsection{Training}
All training was done on a NVIDIA Tesla V100 GPU using
PyTorch \cite{paszke2019pytorch} with 
some operations implemented in CUDA.
All models are trained with the Adam optimizer \cite{kingma2014adam} 
with a learning rate of $0.001$,
using a set of $\numprint{500000}$ points resampled at every epoch with a batch size of 512. 
These points are distributed in a 2:2:1 split of surface, near, and uniform samples. 
We do not make use of positional encodings on the input points.

We train our representation summing together 
the loss functions of the distances at each LOD (see \equref{eq:loss}).
We use $L^2$-distance for our individual per-level losses.
For ShapeNet150 and Thingi32, we train all LODs jointly. 
For TurboSquid16, we use a progressive scheme where we train 
the highest LOD $\maxLOD$ first, and add new trainable levels \linebreak $\lod = \maxLOD-1,\maxLOD-2,\ldots$
every $100$ epochs.
This training scheme slightly benefits lower LODs for more complex shapes.

We briefly experimented with different choices of hyperparameters 
for different architectures (notably for the baselines), but discovered these sets
of hyperparameters worked well across all models. 

\subsection{Rendering}
\label{appendix:rendering}
We implement our baseline renderer
using Python and PyTorch. 
The sparse renderer is implemented using CUDA, cub \cite{merrill2017cub},
and libtorch \cite{paszke2019pytorch}. The implementation takes careful 
advantage of kernel fusion while still making the algorithm agnostic to the architecture. 
The ray-AABB intersection uses Marjercik et. al. \cite{Majercik2018Voxel}.
\secref{sec:octree} provides more details on the sparse octree intersection algorithm.

In the sphere trace, we terminate the algorithm for each individual ray 
if the iteration count
exceeds the maximum or if the stopping criteria ${ \pred{\dist} < \delta }$ 
is reached. We set $\delta = 0.0003$.
In addition, we also check that the step is not oscillating: 
${ |\pred{\dist}_k - \pred{\dist}_{k-1}| < 6 \delta }$ and perform far plane clipping with
depth 5. 
We bound the sphere tracing iterations to $k=200$.

The shadows in the renders are obtained by tracing shadow rays using sphere 
tracing. We also enable SDF ambient occlusion \cite{evans2006fast} and 
materials through matcaps \cite{sloan2001lit}. Surface normals are obtained
using finite differences. 
As noted in the main paper,
the frametimes measured only include the primary ray trace and normal computation
time, and not secondary effects (e.g. shadows). 

\section{Experiment Details}

\subsection{Baselines} 

In this section, we outline the implementation details for DeepSDF \cite{Park19_DeepSDF},
Fourier Feature Network (FFN) \cite{tancik2020fourfeat}, SIREN \cite{sitzmann2019siren}, 
and Neural Implicits \cite{davies2020overfit}. Across all baselines, 
we do not use an activation function at the very last layer to
avoid restrictions on the range of distances the models can output. 
We find this does not significantly affect the results.

\paragraph{DeepSDF} We implement DeepSDF as in the paper,
but remove weight normalization \cite{salimans2016weight}, since we observe
improved performance without it
in our experimental settings. We also do not use latent vectors,
and instead use just the spatial coordinates as input to overfit DeepSDF
to each specific shape.

\paragraph{Fourier Feature Network} We also implement FFN following
the paper, and choose $\sigma = 8$ as it seems to provide the best
overall trade-off between high-frequency noise and detail.
We acknowledge that the reconstruction quality for FFN is very sensitive
to the choice of this hyperparameter; however, we find that it is
time-consuming and therefore impractical to 
search for the optimal $\sigma$ per shape.

\paragraph{SIREN} We implement SIREN following the paper,
and also utilize the weight initialization scheme in the paper.
We do not use the  
the Eikonal regularizer $|\gradient \sdf| = 1$ for our loss function
(and use a simple $L^2$-loss function across all baselines), 
because we find that it is  
important to be able to fit non-metric SDFs that do not satisfy the 
Eikonal equation constraints. Non-metric SDFs are heavily utilized in practice
to make SDF-based content creation easier.

\paragraph{Neural Implicits} We implement Neural Implicits without
any changes to the paper, other than using 
our sampling scheme to generate the dataset so we can control 
training variability across baselines.

\subsection{Reconstruction Metrics}

\vspace{-1em}
\paragraph{Geometry Metrics}
Computing the Chamfer-$L^1$ distance requires surface samples, of both
the ground-truth mesh as well as the predicted SDF.
Typically, these are obtained for the predicted SDF sampling the mesh
extracted with Marching Cubes \cite{lorensen_marching} which introduces
additional error. Instead, we obtain samples by sampling the SDF surface
using ray tracing. 
We uniformly sample $2^{17} = \numprint{131072}$
points in the bounding volume $\boundingVol$,
each assigned with a random spherical direction. We then trace each of these rays using sphere tracing,
and keep adding samples until the minimum number of points are obtained. The stopping criterion is the same as discussed in \ref{appendix:rendering}. We use the 
Chamfer distance as implemented in PyTorch3D \cite{ravi2020pytorch3d}.

\vspace{-2mm}
\paragraph{Image Metrics} We compute the Normal-$L^2$ score by 
sampling 32 evenly distributed, fixed camera positions using a spherical Fibonacci sequence
with radius 4. 
Images are rendered at resolution $512 \times 512$ and surface normals are
evaluated against interpolated surface normals from the reference mesh. 
We evaluate the normal error only on the intersection of the 
predicted and ground-truth masks, since we separately evaluate mask alignment
with intersection over union (iIoU). We use these two metrics because the shape 
silhouettes are perceptually important and surface normals drive the shading.
We use 4 samples per pixel for both images, and implement the mesh 
renderer using Mitsuba 2 \cite{mitsuba2}.

\section{Sparse Ray-Octree Intersection}
\label{sec:octree}

\algblockdefx{ForAll}{EndForAll}[1]%
{\textbf{for all }#1 \textbf{do in parallel}}%
{\textbf{end for}}

\algtext*{EndForAll}

\algdef{SE}[DOWHILE]{Do}{doWhile}{\algorithmicdo}[1]{\algorithmicwhile\ #1}%

We provide more details for the subroutines appearing in Algorithm 1.
Pseudo code for the procedure \textsc{Decide} is listed below:\\

\hrule
\begin{algorithmic}[1]
\Procedure{Decide}{$\mathcal {R}, \mathbf{N}^{(\ell)}, \ell$}
    \ForAll{$t \in \{0,\ldots,|\mathbf{N}^{(\ell)}|-1\}$}
        \State $\{i,j\} \gets\mathbf{N}^{(\ell)}_t$
        \If{$\mathcal{R}_i \cap V^{(\ell)}_j$}
            \If{$\ell = L$}
                \State $\mathbf{D}_t \gets 1$
            \Else
                \State $\mathbf{D}_t \gets \Call{NumChildren}{V^\ell_j}$
            \EndIf
        \Else
            \State $\mathbf{D}_t \gets 0$
        \EndIf
    \EndForAll
    \State \Return{$\mathbf{D}$}
\EndProcedure
\end{algorithmic}
\vspace{1mm}
\hrule
\vspace{5mm}

The \textsc{Decide} procedure determines the voxel-ray pairs that result in intersections. The procedure runs in parallel over (threads) $t$ (line 2). For each $t$, we fetch the ray and voxel indices $i$ and $j$ (line 3). If ray $\mathcal{R}_i$ intersects voxel $V^{(\ell)}_j$ (line 4), we check if we have reached the final level $L$ (line 5). If so, we write a $1$ into list $\mathbf{D}$ at position $t$ (line 6). Otherwise, we write the \textsc{NumChildren} of $V^{(\ell)}_j$ (i.e., the number of occupied children of a voxel in the octree) into list $\mathbf{D}$ at position $t$ (line 8). If ray $\mathcal{R}_i$ does not intersect voxel $V^{(\ell)}_j$, we write $0$ into list $\mathbf{D}$ at position $t$ (line 10). The resulting list $\mathbf{D}$ is returned to the caller (line 11).

Next, we compute the Exclusive Sum of $\mathbf{D}$ and store the resulting list in $\mathbf{S}$. The Exclusive Sum $\mathbf{S}$ of a list of numbers $\mathbf{D}$ is defined as
$$
\mathbf{S}_i = 
\begin{cases} 0 &\mbox{if } i = 0, \\
\sum_{j=0}^{i-1} \mathbf{D}_j & \mbox{otherwise}.
\end{cases}
$$
Note that while this definition appears inherently serial, fast parallel methods for \textsc{ExclusiveSum} are available that treat the problem as a series of parallel reductions \cite{Blelloch1990, harris2007parallel}. 
The exclusive sum is a powerful parallel programming construct that provides the index for writing data into a list from independent threads without conflicts (write hazards).

This can be seen in the pseudo code for procedure \textsc{Compactify} called at the final step of iteration in Algorithm 1:\\

\hrule
\begin{algorithmic}[1]
\Procedure{Compactify}{$\mathbf{N}^{(\ell)}, \mathbf{D}, \mathbf{S}$}
    \ForAll{$t \in \{0,\ldots,|\mathbf{N}^{(\ell)}|-1\}$}
        \If{$\mathbf{D}_t = 1$}
            \State $k \gets \mathbf{S}_t$
            \State $\mathbf{N}^{(\ell+1)}_k \gets \mathbf{N}^{(\ell)}_t$
        \EndIf
    \EndForAll
    \State \Return{$\mathbf{N}^{(\ell+1)}$}
\EndProcedure
\end{algorithmic}
\vspace{1mm}
\hrule
\vspace{1mm}
The \textsc{Compactify} subroutine removes all ray-voxel pairs that do not result in an intersection (and thus do not contribute to $\mathbf{S}$). This routine is run in parallel over $t$ (line 2). When $\mathbf{D}_t = 1$, meaning voxel $\mathcal{V}^\ell_t$ was hit (line 3), we copy the ray/voxel index pair from $\mathbf{N}^{(\ell)}_t$ to its new location $k$ obtained from the exclusive sum result $\mathbf{S}_t$ (line 4), $\mathbf{N}^{(\ell+1)}$ (line 5). We then return the new list $\mathbf{N}^{(\ell+1)}$ to the caller.

If the iteration has not reached the final step, i.e. $l\ne L$ in Algorithm 1, we call \textsc{Subdivide} listed below:\\

\hrule
\begin{algorithmic}[1]
\Procedure{Subdivide}{$\mathbf{N}^{(\ell)}, \mathbf{D}, \mathbf{S}$}
    \ForAll{$t \in \{0,\ldots,|\mathbf{N}^\setlength{}{}{(\ell)}|-1\}$}
        \If{$\mathbf{D}_t \ne 0$}
            \State $\{i,j\} \gets\mathbf{N}^{(\ell)}_t$
            \State $k \gets \mathbf{S}_t$
            \For{$c \in \Call{OrderedChildren}{\mathcal{R}_i, V^{(\ell)}_j}$}
                \State $\mathbf{N}^{(\ell+1)}_k \gets \{i, c\}$
                \State $k \gets k+1$
            \EndFor
        \EndIf
    \EndForAll
    \State \Return{$\mathbf{N}^{\ell+1}$}
\EndProcedure
\end{algorithmic}
\vspace{1mm}
\hrule
\vspace{3mm}
The \textsc{Subdivide} populates the next list $\mathbf{N}^{(\ell+1)}$ by subdividing out $\mathbf{N}^{(\ell)}$. This routine is run in parallel over $t$ (line 2). 
When $\mathbf{D}_t \ne 0$, meaning voxel $\mathcal{V}^{(\ell)}_t$ was hit (line 3), we do the following: We load the ray/voxel index pair $\{i,j\}$ from $\mathbf{N}^\ell_t$ (line 4). The output index $k$ for the first child voxel index is obtained (line 5). We then iterate over the ordered children of the current voxel $V^{(\ell)}_j$ using iterator \textsc{OrderedChildren} (line 6). This iterator returns the child voxels of $V^{(\ell)}_j$ in front-to-back order with respect to ray $\mathcal{R}_i$. This ordering is only dependant on which of the 8 octants of space contains the origin of the ray, and can be stored in a pre-computed $8\times8$ table. We write the ray/voxel index pair to the new list $\mathbf{N}^{(\ell+1)}$ at position $k$ (line 7). The output index $k$ is incremented (line 8), and the resulting list of (subdivided) ray/voxel index pairs (line 9).

\section{Additional Results}

More result examples from each dataset used can be found in the following pages. We also refer to our supplementary video for a real-time demonstration of our method.

\input{figures/supp_ts_comp}
\input{figures/supp_thingi_comp}
\input{figures/supp_sn_comp}

\section{Artist Acknowledgements}

We credit the following artists for the 3D assets used in this work. In alphabetical order:
3D Aries (\textit{Cogs}), abramsdesign (\textit{Cabin}), the Art Institute of Chicago (\textit{Lion}), Distefan (\textit{Train}), DRONNNNN95 (\textit{House}), Dmitriev Vasiliy (\textit{V Mech}), Felipe Alfonso (\textit{Cheese}), Florian Berger (\textit{Oldcar}), Gary Warne (\textit{Mobius}), Inigo Quilez (\textit{Snail}), klk (\textit{Teapot}), Martijn Steinrucken (\textit{Snake}), Max 3D Design (\textit{Robot}), monsterkodi (\textit{Skull}), QE3D (\textit{Parthenon}), RaveeCG (\textit{Horseman}), sam\_rus (\textit{City}), the Stanford Computer Graphics Lab (\textit{Lucy}), TheDizajn (\textit{Boat}), Xor (\textit{Burger}, \textit{Fish}), your artist (\textit{Chameleon}), and zames1992 (\textit{Cathedral}).

%% file: figures/supp_ts_comp.tex

\begin{figure*}[t]
    \centering
    \vspace{-5pt}
    \begin{subfigure}[t]{0.2\linewidth}
        \centering
        \includegraphics[width=\textwidth]{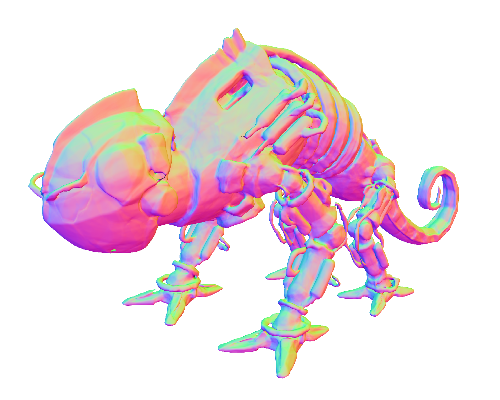}\vspace{-5pt}
        \caption*{DeepSDF \cite{Park19_DeepSDF}}
    \end{subfigure}%
    \begin{subfigure}[t]{0.2\linewidth}
        \centering
        \includegraphics[width=\textwidth]{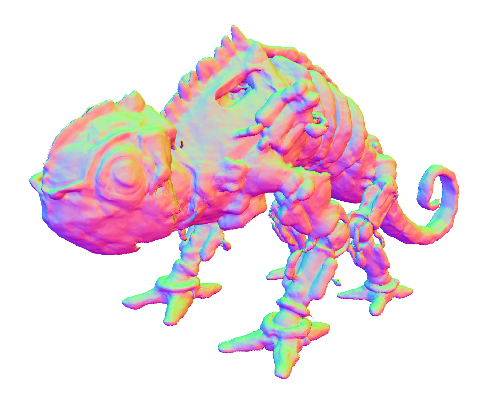}\vspace{-5pt}
        \caption*{FFN \cite{tancik2020fourfeat}}
    \end{subfigure}%
    \begin{subfigure}[t]{0.2\linewidth}
        \centering
        \includegraphics[width=\textwidth]{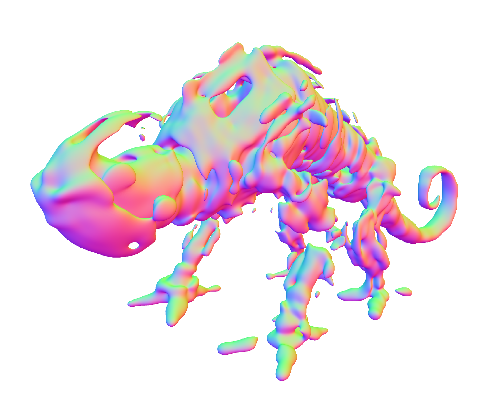}\vspace{-5pt}
        \caption*{SIREN \cite{sitzmann2019siren}}
    \end{subfigure}%
    \begin{subfigure}[t]{0.2\linewidth}
        \centering
        \includegraphics[width=\textwidth]{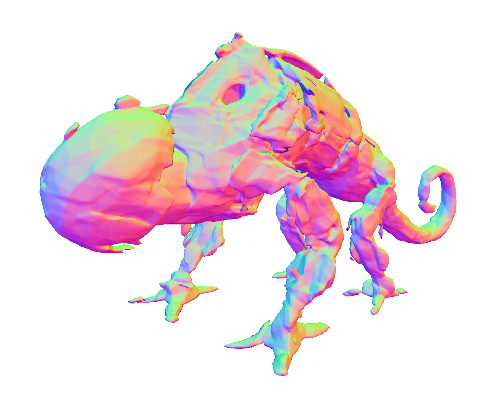}\vspace{-5pt}
        \caption*{Neural Implicits \cite{davies2020overfit}}
    \end{subfigure}%
    \begin{subfigure}[t]{0.2\linewidth}
        \centering
        \includegraphics[width=\textwidth]{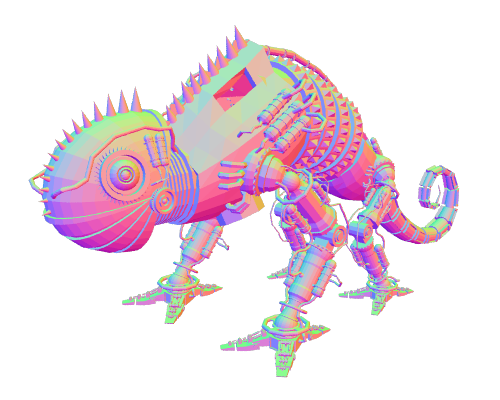}\vspace{-5pt}
        \caption*{Reference}
    \end{subfigure}
    \begin{subfigure}[t]{0.2\linewidth}
        \centering
        \includegraphics[width=\textwidth]{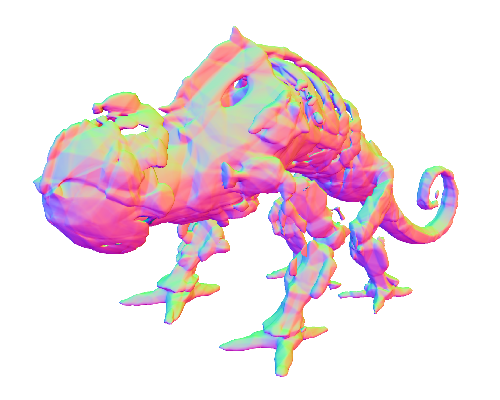}\vspace{-5pt}
        \caption*{Ours / LOD 2}
    \end{subfigure}%
    \begin{subfigure}[t]{0.2\linewidth}
        \centering
        \includegraphics[width=\textwidth]{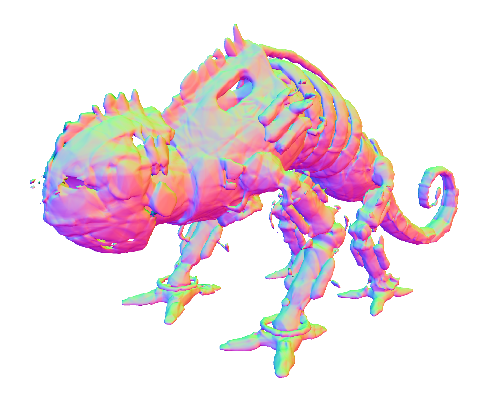}\vspace{-5pt}
        \caption*{Ours / LOD 3}
    \end{subfigure}%
    \begin{subfigure}[t]{0.2\linewidth}
        \centering
        \includegraphics[width=\textwidth]{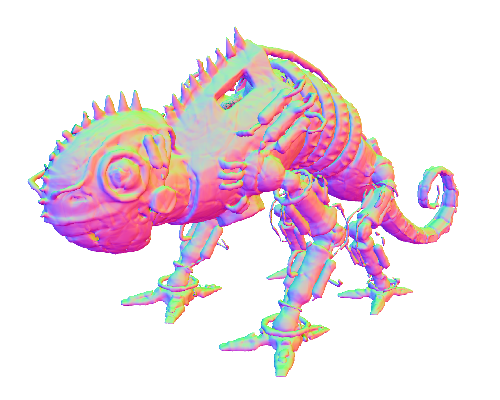}\vspace{-5pt}
        \caption*{Ours / LOD 4}
    \end{subfigure}%
    \begin{subfigure}[t]{0.2\linewidth}
        \centering
        \includegraphics[width=\textwidth]{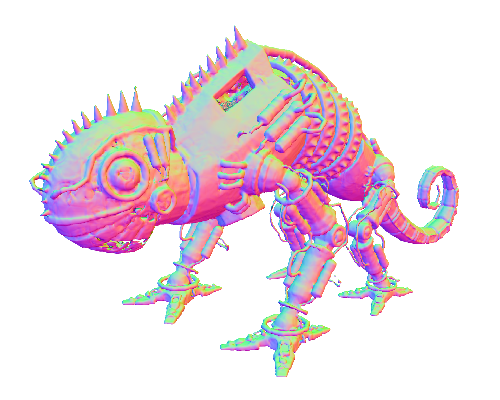}\vspace{-5pt}
        \caption*{Ours / LOD 5}
    \end{subfigure}%
    \begin{subfigure}[t]{0.2\linewidth}
        \centering
        \includegraphics[width=\textwidth]{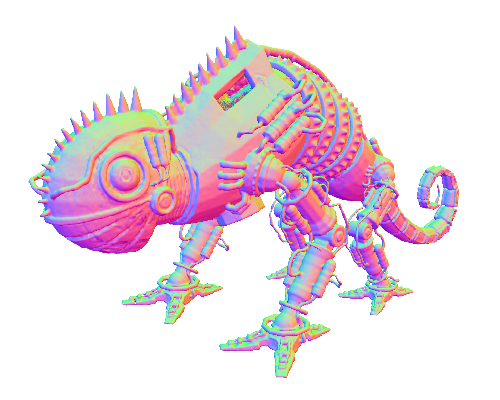}\vspace{-5pt}
        \caption*{Ours / LOD 6}
    \end{subfigure}
    \begin{subfigure}[t]{0.2\linewidth}
        \centering
        \includegraphics[width=\textwidth]{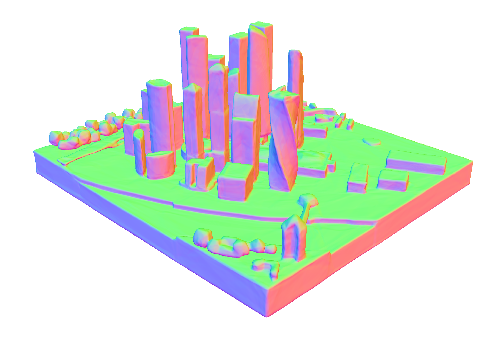}\vspace{-8pt}
        \caption*{DeepSDF \cite{Park19_DeepSDF}}
    \end{subfigure}%
    \begin{subfigure}[t]{0.2\linewidth}
        \centering
        \includegraphics[width=\textwidth]{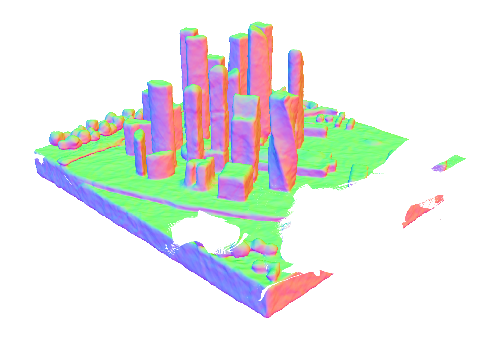}\vspace{-8pt}
        \caption*{FFN \cite{tancik2020fourfeat}}
    \end{subfigure}%
    \begin{subfigure}[t]{0.2\linewidth}
        \centering
        \includegraphics[width=\textwidth]{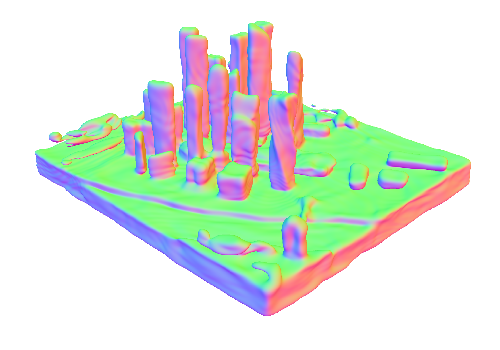}\vspace{-8pt}
        \caption*{SIREN \cite{sitzmann2019siren}}
    \end{subfigure}%
    \begin{subfigure}[t]{0.2\linewidth}
        \centering
        \includegraphics[width=\textwidth]{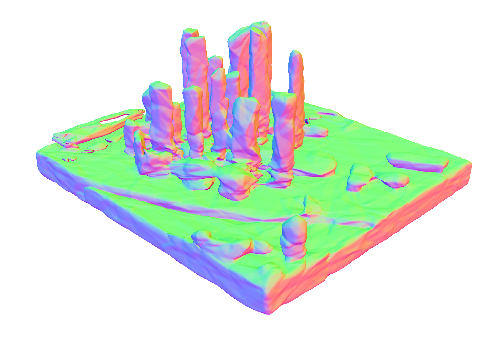}\vspace{-8pt}
        \caption*{Neural Implicits \cite{davies2020overfit}}
    \end{subfigure}%
    \begin{subfigure}[t]{0.2\linewidth}
        \centering
        \includegraphics[width=\textwidth]{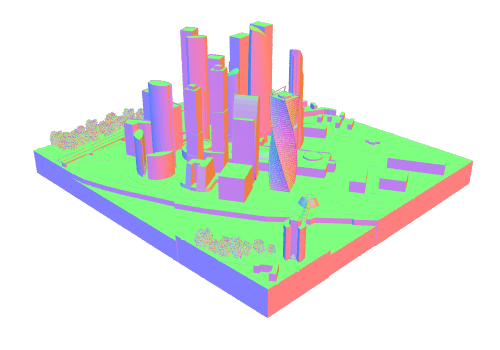}\vspace{-8pt}
        \caption*{Reference}
    \end{subfigure}
    \begin{subfigure}[t]{0.2\linewidth}
        \centering
        \includegraphics[width=\textwidth]{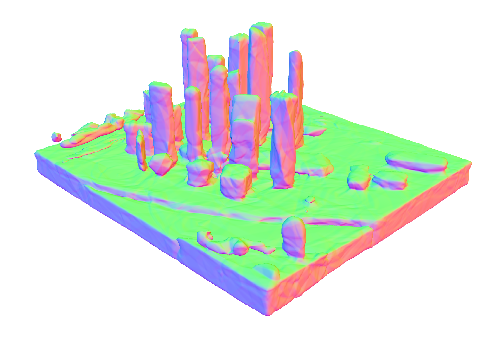}\vspace{-8pt}
        \caption*{Ours / LOD 2}
    \end{subfigure}%
    \begin{subfigure}[t]{0.2\linewidth}
        \centering
        \includegraphics[width=\textwidth]{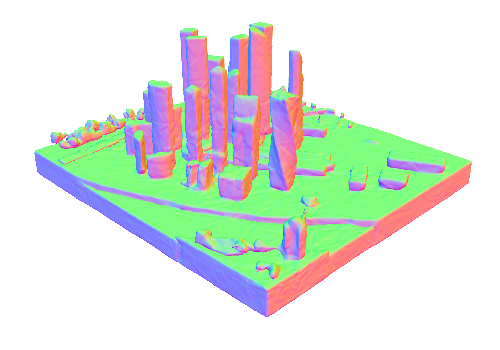}\vspace{-8pt}
        \caption*{Ours / LOD 3}
    \end{subfigure}%
    \begin{subfigure}[t]{0.2\linewidth}
        \centering
        \includegraphics[width=\textwidth]{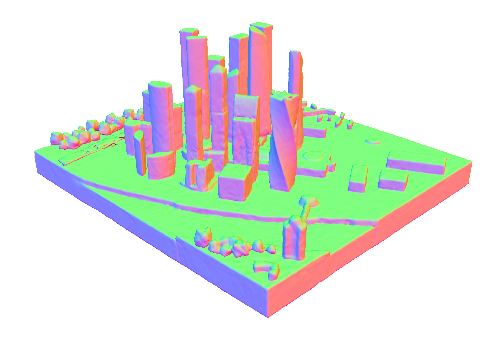}\vspace{-8pt}
        \caption*{Ours / LOD 4}
    \end{subfigure}%
    \begin{subfigure}[t]{0.2\linewidth}
        \centering
        \includegraphics[width=\textwidth]{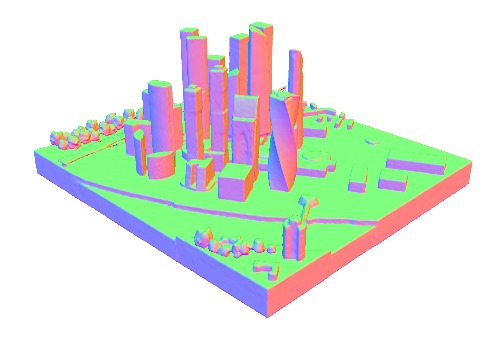}\vspace{-8pt}
        \caption*{Ours / LOD 5}
    \end{subfigure}%
        \begin{subfigure}[t]{0.2\linewidth}
        \centering
        \includegraphics[width=\textwidth]{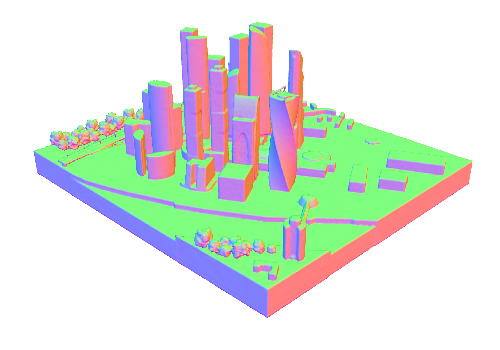}\vspace{-8pt}
        \caption*{Ours / LOD 6}
    \end{subfigure}
    \begin{subfigure}[t]{0.2\linewidth}
        \centering
        \includegraphics[width=\textwidth]{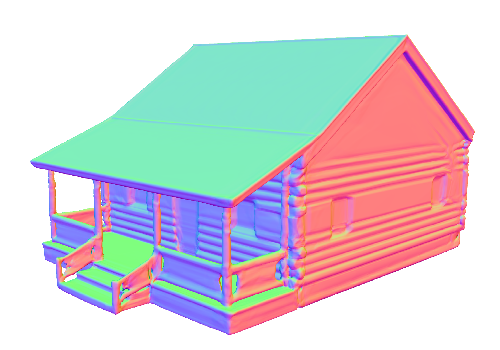}\vspace{-6pt}
        \caption*{DeepSDF \cite{Park19_DeepSDF}}
    \end{subfigure}%
    \begin{subfigure}[t]{0.2\linewidth}
        \centering
        \includegraphics[width=\textwidth]{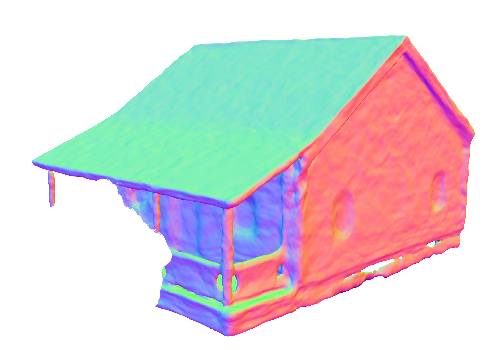}\vspace{-6pt}
        \caption*{FFN \cite{tancik2020fourfeat}}
    \end{subfigure}%
    \begin{subfigure}[t]{0.2\linewidth}
        \centering
        \includegraphics[width=\textwidth]{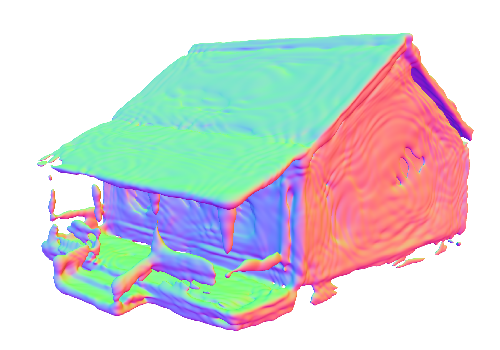}\vspace{-6pt}
        \caption*{SIREN \cite{sitzmann2019siren}}
    \end{subfigure}%
    \begin{subfigure}[t]{0.2\linewidth}
        \centering
        \includegraphics[width=\textwidth]{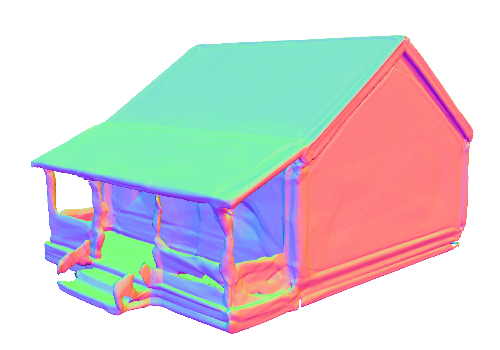}\vspace{-6pt}
        \caption*{Neural Implicits \cite{davies2020overfit}}
    \end{subfigure}%
    \begin{subfigure}[t]{0.2\linewidth}
        \centering
        \includegraphics[width=\textwidth]{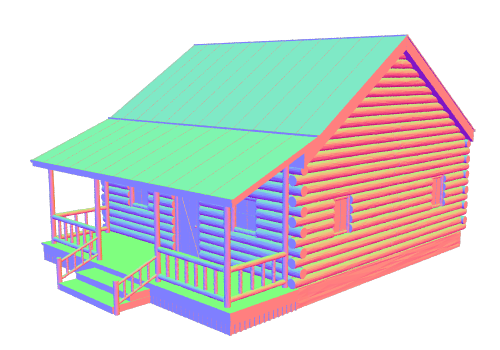}\vspace{-6pt}
        \caption*{Reference}
    \end{subfigure}
    \begin{subfigure}[t]{0.2\linewidth}
        \centering
        \includegraphics[width=\textwidth]{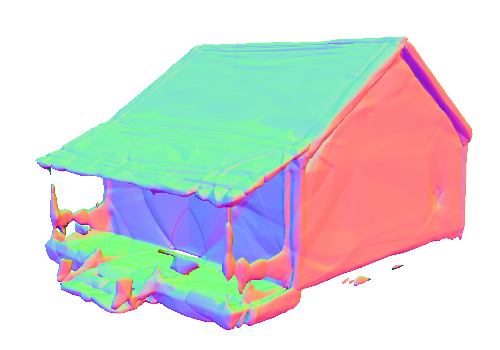}\vspace{-6pt}
        \caption*{Ours / LOD 1}
    \end{subfigure}%
    \begin{subfigure}[t]{0.2\linewidth}
        \centering
        \includegraphics[width=\textwidth]{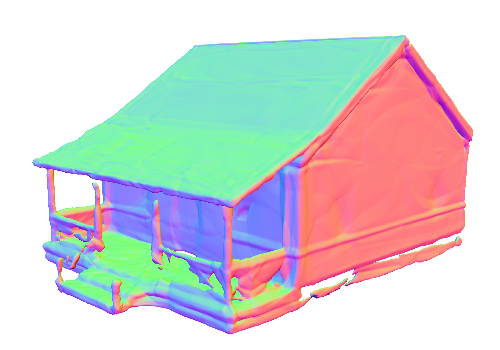}\vspace{-6pt}
        \caption*{Ours / LOD 2}
    \end{subfigure}%
    \begin{subfigure}[t]{0.2\linewidth}
        \centering
        \includegraphics[width=\textwidth]{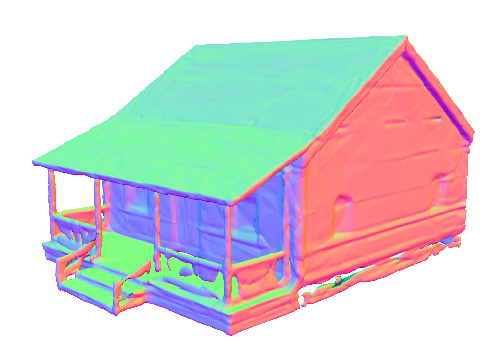}\vspace{-6pt}
        \caption*{Ours / LOD 3}
    \end{subfigure}%
    \begin{subfigure}[t]{0.2\linewidth}
        \centering
        \includegraphics[width=\textwidth]{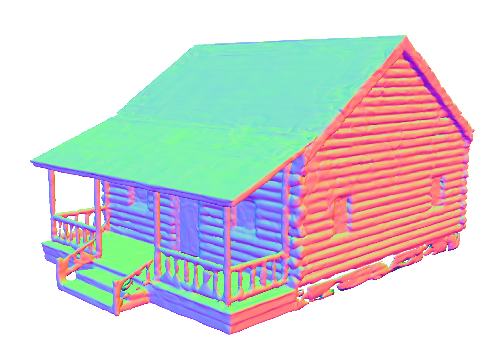}\vspace{-6pt}
        \caption*{Ours / LOD 4}
    \end{subfigure}%
    \begin{subfigure}[t]{0.2\linewidth}
        \centering
        \includegraphics[width=\textwidth]{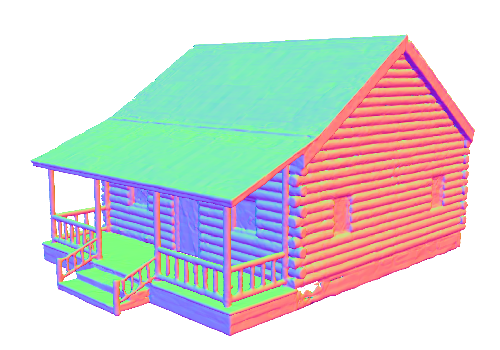}\vspace{-6pt}
        \caption*{Ours / LOD 5}
    \end{subfigure}%
    \caption{\textbf{Additional TurboSquid16 Results.} FFN exhibits white patch artifacts (e.g. City and Cabin) because it struggles to
    learn a conservative metric SDF, resulting in the sphere tracing algorithm missing the surface entirely. Best viewed zoomed in.}
\label{fig:supp_ts_comp}
\vspace{-9pt}
\end{figure*}

%% file: figures/supp_thingi_comp.tex

\begin{figure*}[t]
    \centering
    \vspace{-5pt}
    \begin{subfigure}[t]{0.2\linewidth}
        \centering
        \includegraphics[width=\textwidth]{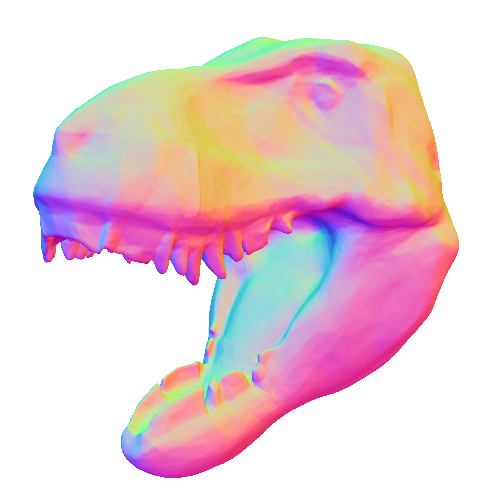}\vspace{-5pt}
        \caption*{DeepSDF \cite{Park19_DeepSDF}}
    \end{subfigure}%
    \begin{subfigure}[t]{0.2\linewidth}
        \centering
        \includegraphics[width=\textwidth]{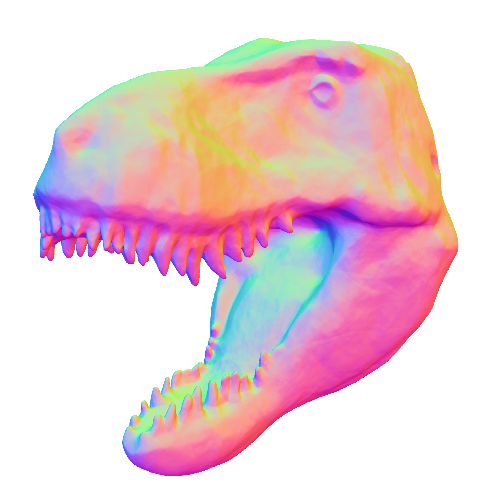}\vspace{-5pt}
        \caption*{FFN \cite{tancik2020fourfeat}}
    \end{subfigure}%
    \begin{subfigure}[t]{0.2\linewidth}
        \centering
        \includegraphics[width=\textwidth]{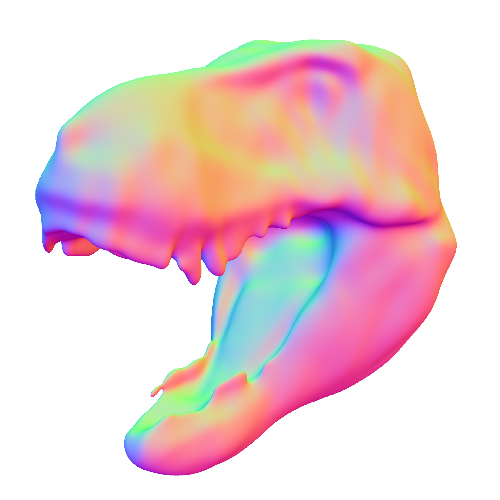}\vspace{-5pt}
        \caption*{SIREN \cite{sitzmann2019siren}}
    \end{subfigure}%
    \begin{subfigure}[t]{0.2\linewidth}
        \centering
        \includegraphics[width=\textwidth]{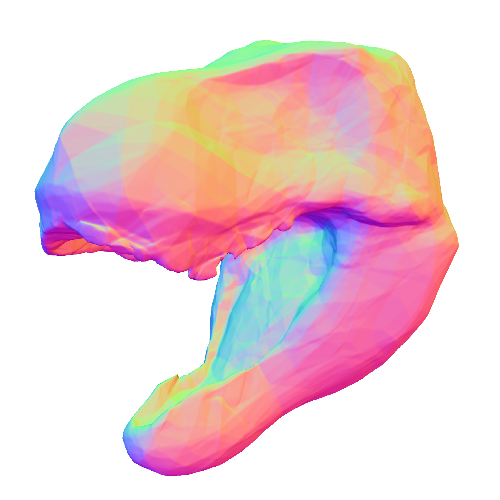}\vspace{-5pt}
        \caption*{Neural Implicits \cite{davies2020overfit}}
    \end{subfigure}%
    \begin{subfigure}[t]{0.2\linewidth}
        \centering
        \includegraphics[width=\textwidth]{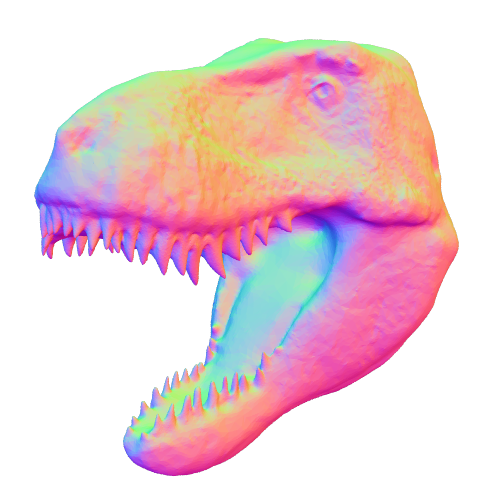}\vspace{-5pt}
        \caption*{Reference}
    \end{subfigure}
    \begin{subfigure}[t]{0.2\linewidth}
        \centering
        \includegraphics[width=\textwidth]{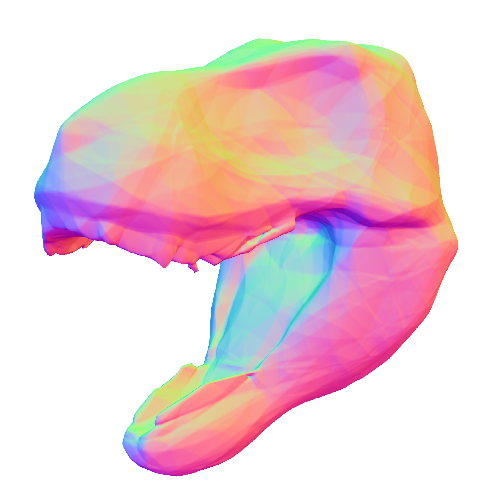}\vspace{-5pt}
        \caption*{Ours / LOD 1}
    \end{subfigure}%
    \begin{subfigure}[t]{0.2\linewidth}
        \centering
        \includegraphics[width=\textwidth]{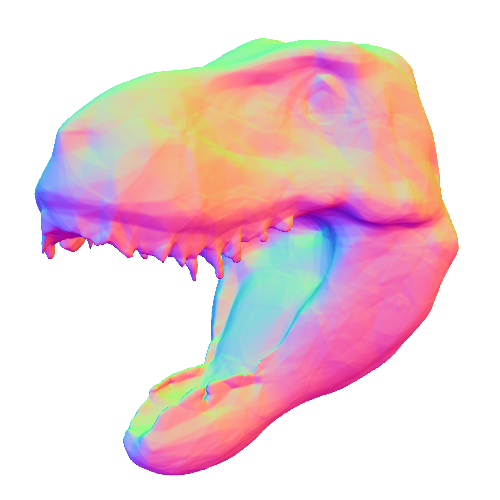}\vspace{-5pt}
        \caption*{Ours / LOD 2}
    \end{subfigure}%
    \begin{subfigure}[t]{0.2\linewidth}
        \centering
        \includegraphics[width=\textwidth]{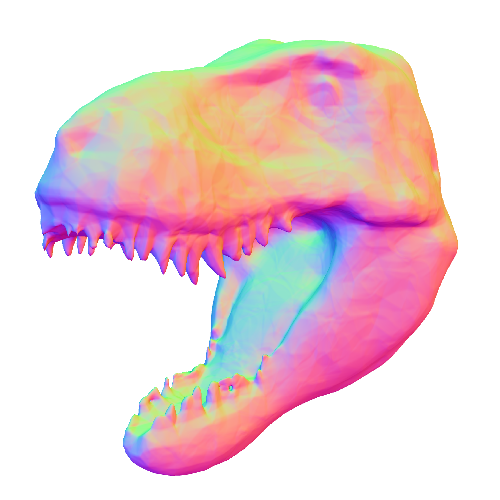}\vspace{-5pt}
        \caption*{Ours / LOD 3}
    \end{subfigure}%
    \begin{subfigure}[t]{0.2\linewidth}
        \centering
        \includegraphics[width=\textwidth]{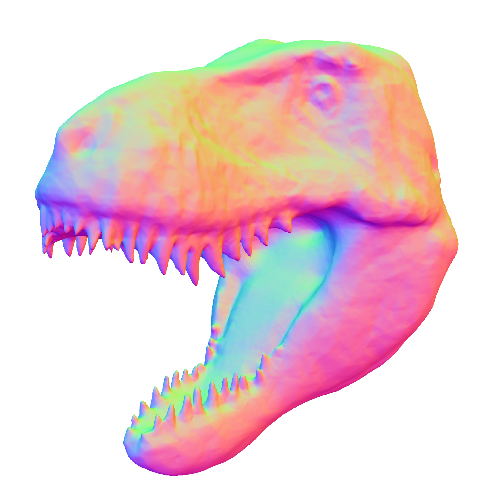}\vspace{-5pt}
        \caption*{Ours / LOD 4}
    \end{subfigure}%
    \begin{subfigure}[t]{0.2\linewidth}
        \centering
        \includegraphics[width=\textwidth]{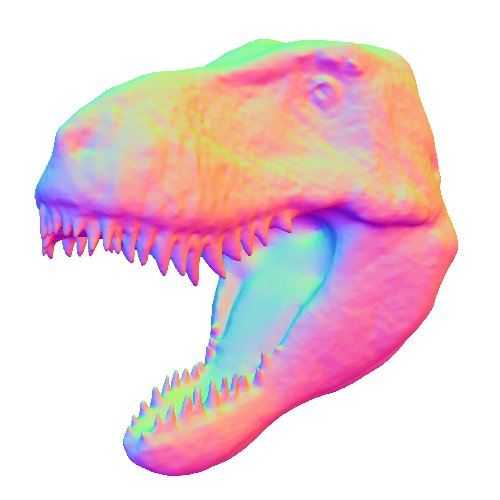}\vspace{-5pt}
        \caption*{Ours / LOD 5}
    \end{subfigure}
    \begin{subfigure}[t]{0.2\linewidth}
        \centering
        \includegraphics[width=\textwidth]{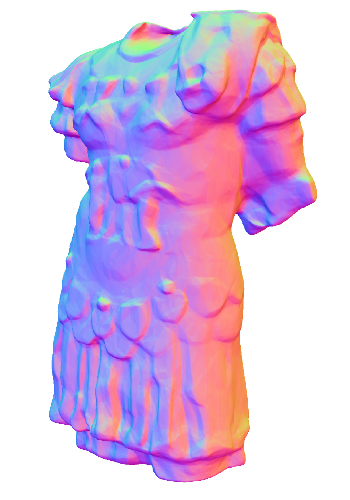}\vspace{-5pt}
        \caption*{DeepSDF \cite{Park19_DeepSDF}}
    \end{subfigure}%
    \begin{subfigure}[t]{0.2\linewidth}
        \centering
        \includegraphics[width=\textwidth]{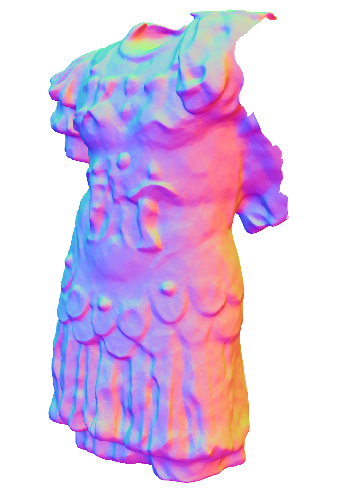}\vspace{-5pt}
        \caption*{FFN \cite{tancik2020fourfeat}}
    \end{subfigure}%
    \begin{subfigure}[t]{0.2\linewidth}
        \centering
        \includegraphics[width=\textwidth]{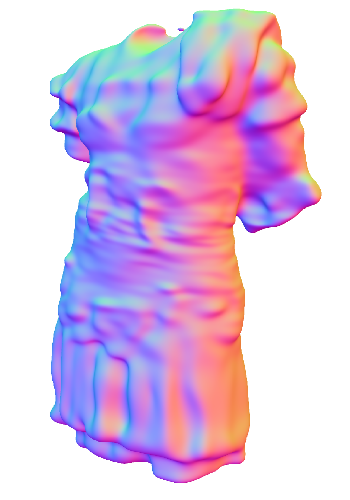}\vspace{-5pt}
        \caption*{SIREN \cite{sitzmann2019siren}}
    \end{subfigure}%
    \begin{subfigure}[t]{0.2\linewidth}
        \centering
        \includegraphics[width=\textwidth]{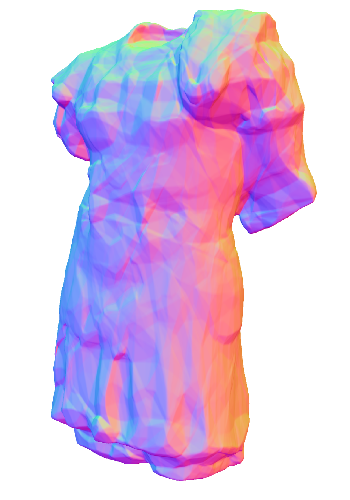}\vspace{-5pt}
        \caption*{Neural Implicits \cite{davies2020overfit}}
    \end{subfigure}%
    \begin{subfigure}[t]{0.2\linewidth}
        \centering
        \includegraphics[width=\textwidth]{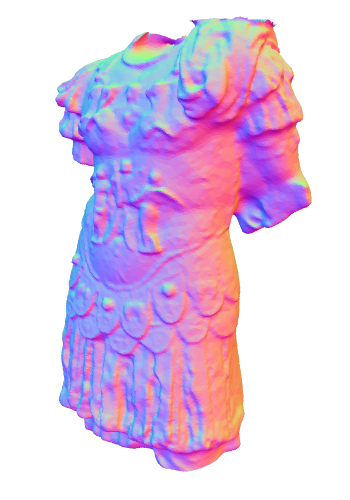}\vspace{-5pt}
        \caption*{Reference}
    \end{subfigure}
    \begin{subfigure}[t]{0.2\linewidth}
        \centering
        \includegraphics[width=\textwidth]{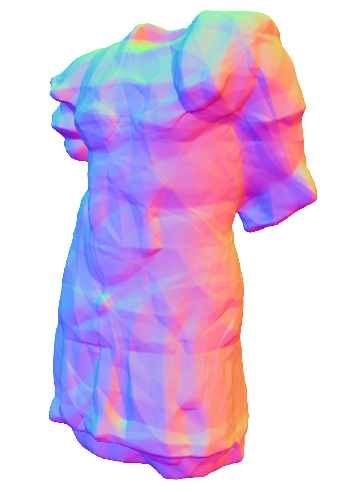}\vspace{-5pt}
        \caption*{Ours / LOD 1}
    \end{subfigure}%
    \begin{subfigure}[t]{0.2\linewidth}
        \centering
        \includegraphics[width=\textwidth]{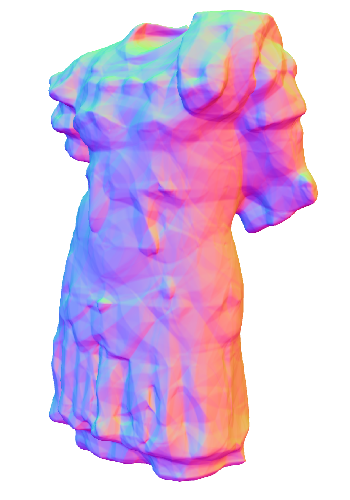}\vspace{-5pt}
        \caption*{Ours / LOD 2}
    \end{subfigure}%
    \begin{subfigure}[t]{0.2\linewidth}
        \centering
        \includegraphics[width=\textwidth]{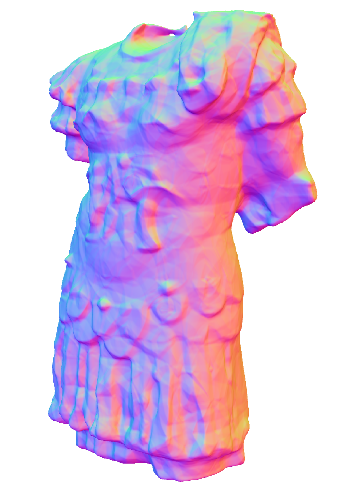}\vspace{-5pt}
        \caption*{Ours / LOD 3}
    \end{subfigure}%
    \begin{subfigure}[t]{0.2\linewidth}
        \centering
        \includegraphics[width=\textwidth]{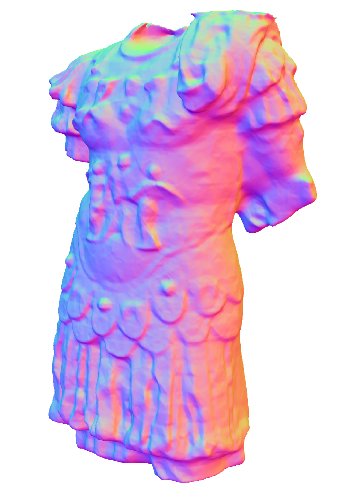}\vspace{-5pt}
        \caption*{Ours / LOD 4}
    \end{subfigure}%
    \begin{subfigure}[t]{0.2\linewidth}
        \centering
        \includegraphics[width=\textwidth]{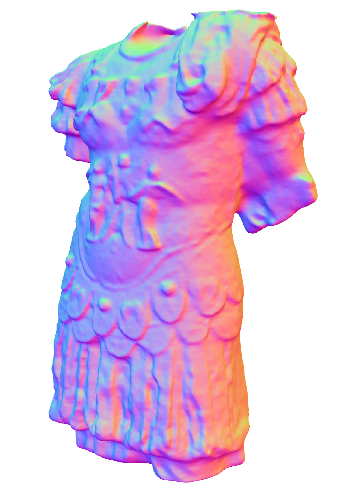}\vspace{-5pt}
        \caption*{Ours / LOD 5}
    \end{subfigure}%
    \caption{\textbf{Additional Thingi32 Results.} Best viewed zoomed in.}
\label{fig:supp_thingi_comp}
\vspace{-9pt}
\end{figure*}

%% file: figures/supp_sn_comp.tex

\begin{figure*}[t]
    \centering
    \vspace{-5pt}
    \begin{subfigure}[t]{0.2\linewidth}
        \centering
        \includegraphics[width=\textwidth]{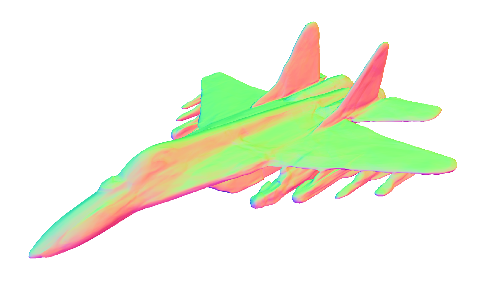}\vspace{-5pt}
        \caption*{DeepSDF \cite{Park19_DeepSDF}}
    \end{subfigure}%
    \begin{subfigure}[t]{0.2\linewidth}
        \centering
        \includegraphics[width=\textwidth]{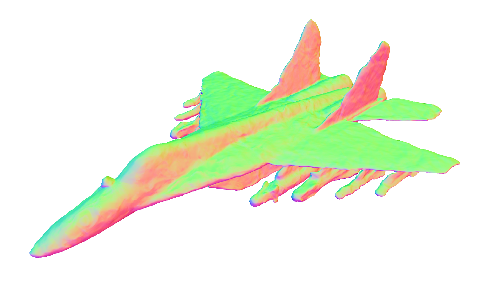}\vspace{-5pt}
        \caption*{FFN \cite{tancik2020fourfeat}}
    \end{subfigure}%
    \begin{subfigure}[t]{0.2\linewidth}
        \centering
        \includegraphics[width=\textwidth]{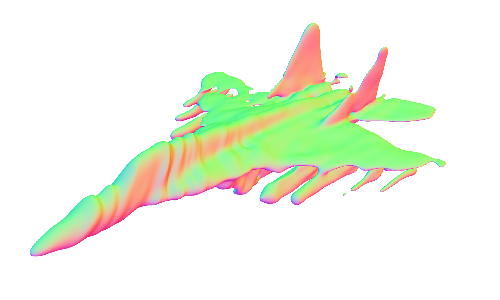}\vspace{-5pt}
        \caption*{SIREN \cite{sitzmann2019siren}}
    \end{subfigure}%
    \begin{subfigure}[t]{0.2\linewidth}
        \centering
        \includegraphics[width=\textwidth]{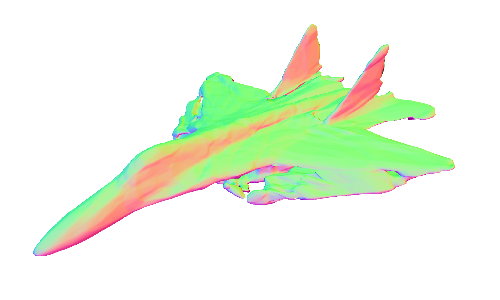}\vspace{-5pt}
        \caption*{Neural Implicits \cite{davies2020overfit}}
    \end{subfigure}%
    \begin{subfigure}[t]{0.2\linewidth}
        \centering
        \includegraphics[width=\textwidth]{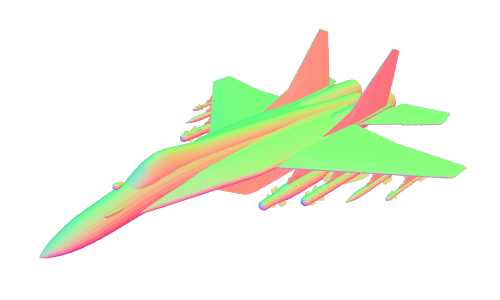}\vspace{-5pt}
        \caption*{Reference}
    \end{subfigure}
    \begin{subfigure}[t]{0.2\linewidth}
        \centering
        \includegraphics[width=\textwidth]{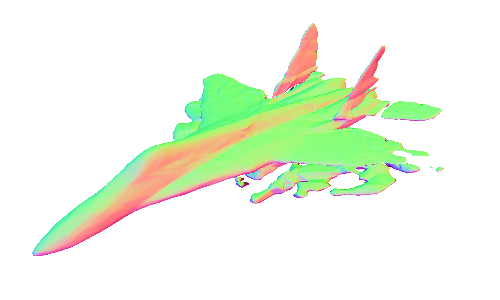}\vspace{-5pt}
        \caption*{Ours / LOD 1}
    \end{subfigure}%
    \begin{subfigure}[t]{0.2\linewidth}
        \centering
        \includegraphics[width=\textwidth]{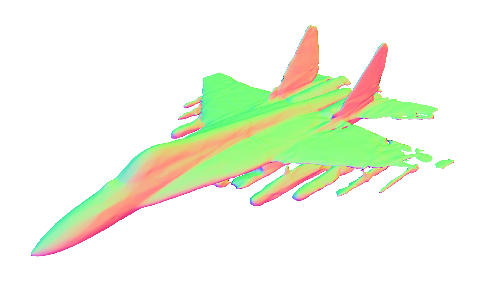}\vspace{-5pt}
        \caption*{Ours / LOD 2}
    \end{subfigure}%
    \begin{subfigure}[t]{0.2\linewidth}
        \centering
        \includegraphics[width=\textwidth]{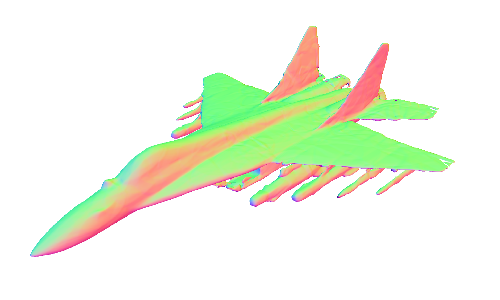}\vspace{-5pt}
        \caption*{Ours / LOD 3}
    \end{subfigure}%
    \begin{subfigure}[t]{0.2\linewidth}
        \centering
        \includegraphics[width=\textwidth]{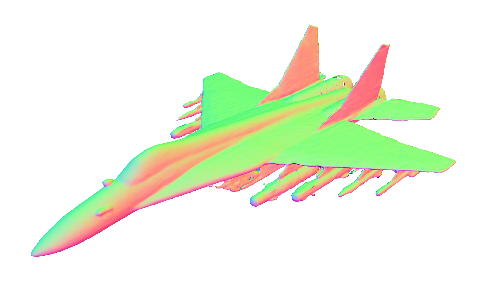}\vspace{-5pt}
        \caption*{Ours / LOD 4}
    \end{subfigure}%
    \begin{subfigure}[t]{0.2\linewidth}
        \centering
        \includegraphics[width=\textwidth]{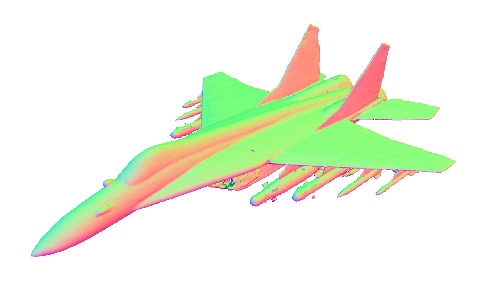}\vspace{-5pt}
        \caption*{Ours / LOD 5}
    \end{subfigure}
    \begin{subfigure}[t]{0.2\linewidth}
        \centering
        \includegraphics[width=\textwidth]{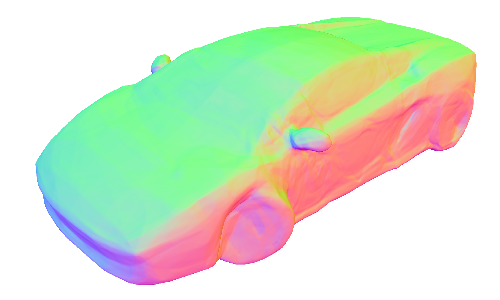}\vspace{-5pt}
        \caption*{DeepSDF \cite{Park19_DeepSDF}}
    \end{subfigure}%
    \begin{subfigure}[t]{0.2\linewidth}
        \centering
        \includegraphics[width=\textwidth]{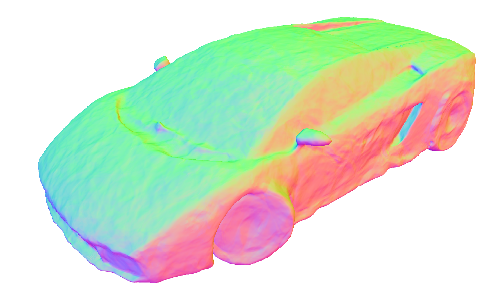}\vspace{-5pt}
        \caption*{FFN \cite{tancik2020fourfeat}}
    \end{subfigure}%
    \begin{subfigure}[t]{0.2\linewidth}
        \centering
        \includegraphics[width=\textwidth]{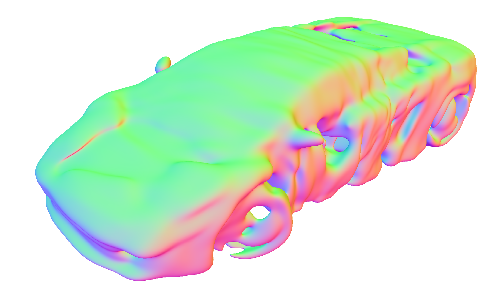}\vspace{-5pt}
        \caption*{SIREN \cite{sitzmann2019siren}}
    \end{subfigure}%
    \begin{subfigure}[t]{0.2\linewidth}
        \centering
        \includegraphics[width=\textwidth]{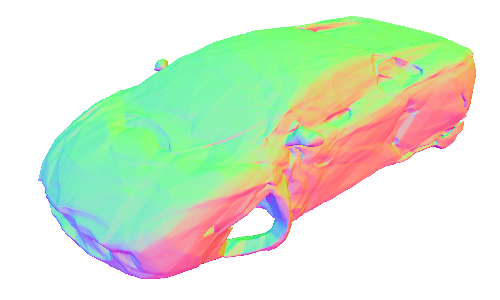}\vspace{-5pt}
        \caption*{Neural Implicits \cite{davies2020overfit}}
    \end{subfigure}%
    \begin{subfigure}[t]{0.2\linewidth}
        \centering
        \includegraphics[width=\textwidth]{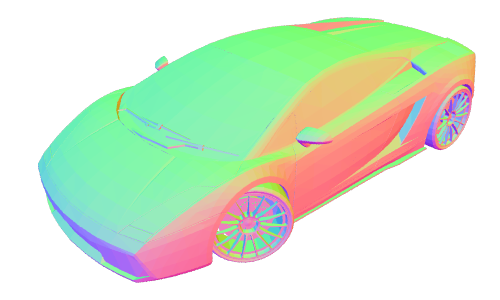}\vspace{-5pt}
        \caption*{Reference}
    \end{subfigure}
    \begin{subfigure}[t]{0.2\linewidth}
        \centering
        \includegraphics[width=\textwidth]{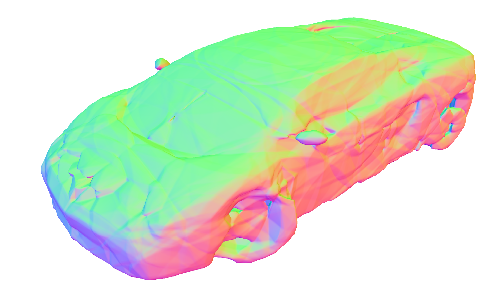}\vspace{-5pt}
        \caption*{Ours / LOD 1}
    \end{subfigure}%
    \begin{subfigure}[t]{0.2\linewidth}
        \centering
        \includegraphics[width=\textwidth]{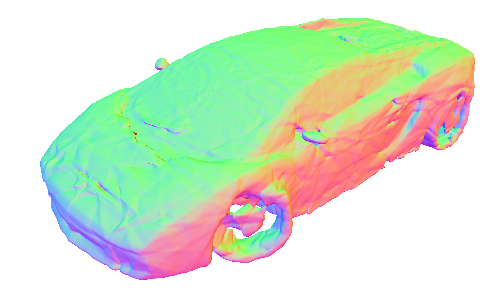}\vspace{-5pt}
        \caption*{Ours / LOD 2}
    \end{subfigure}%
    \begin{subfigure}[t]{0.2\linewidth}
        \centering
        \includegraphics[width=\textwidth]{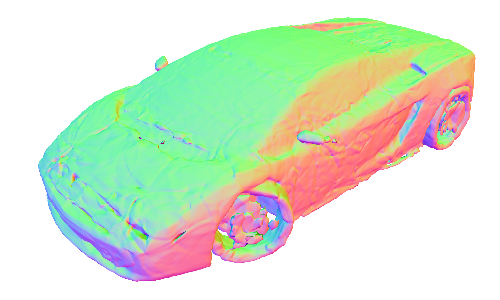}\vspace{-5pt}
        \caption*{Ours / LOD 3}
    \end{subfigure}%
    \begin{subfigure}[t]{0.2\linewidth}
        \centering
        \includegraphics[width=\textwidth]{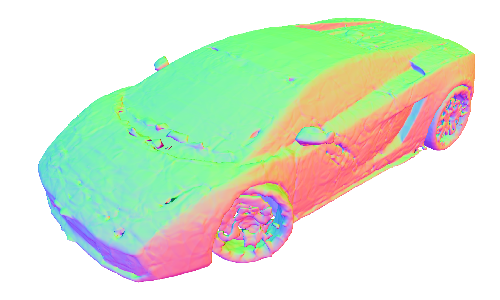}\vspace{-5pt}
        \caption*{Ours / LOD 4}
    \end{subfigure}%
    \begin{subfigure}[t]{0.2\linewidth}
        \centering
        \includegraphics[width=\textwidth]{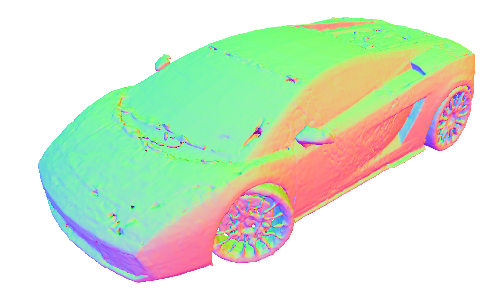}\vspace{-5pt}
        \caption*{Ours / LOD 5}
    \end{subfigure}

    \begin{subfigure}[t]{0.2\linewidth}
        \centering
        \includegraphics[width=\textwidth]{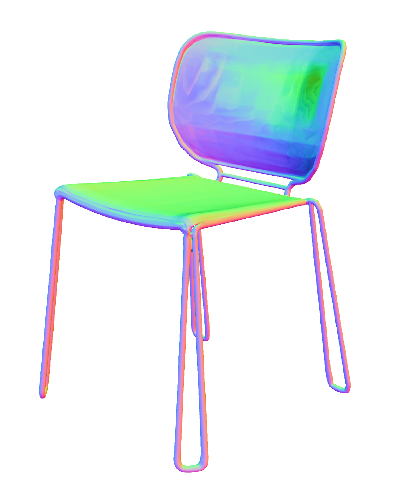}\vspace{-5pt}
        \caption*{DeepSDF \cite{Park19_DeepSDF}}
    \end{subfigure}%
    \begin{subfigure}[t]{0.2\linewidth}
        \centering
        \includegraphics[width=\textwidth]{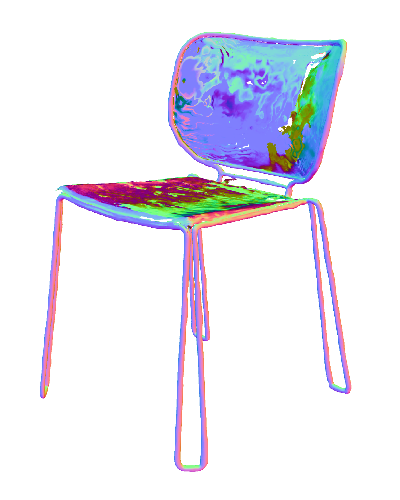}\vspace{-5pt}
        \caption*{FFN \cite{tancik2020fourfeat}}
    \end{subfigure}%
    \begin{subfigure}[t]{0.2\linewidth}
        \centering
        \includegraphics[width=\textwidth]{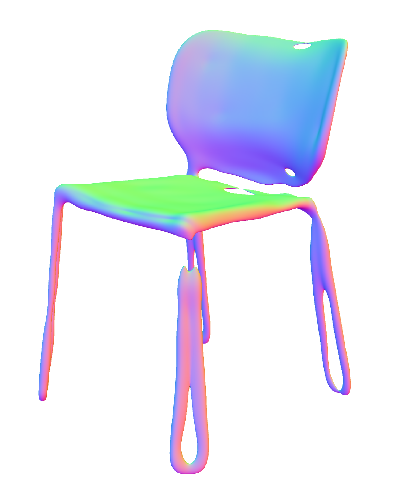}\vspace{-5pt}
        \caption*{SIREN \cite{sitzmann2019siren}}
    \end{subfigure}%
    \begin{subfigure}[t]{0.2\linewidth}
        \centering
        \includegraphics[width=\textwidth]{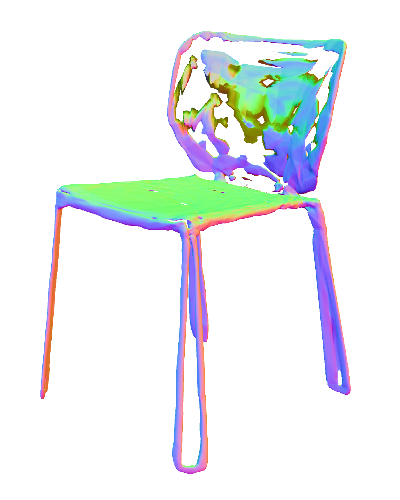}\vspace{-5pt}
        \caption*{Neural Implicits \cite{davies2020overfit}}
    \end{subfigure}%
    \begin{subfigure}[t]{0.2\linewidth}
        \centering
        \includegraphics[width=\textwidth]{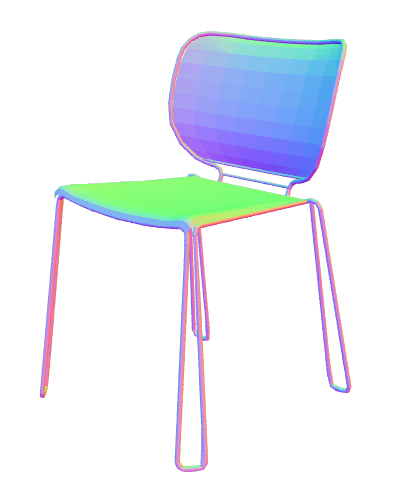}\vspace{-5pt}
        \caption*{Reference}
    \end{subfigure}
    \begin{subfigure}[t]{0.2\linewidth}
        \centering
        \includegraphics[width=\textwidth]{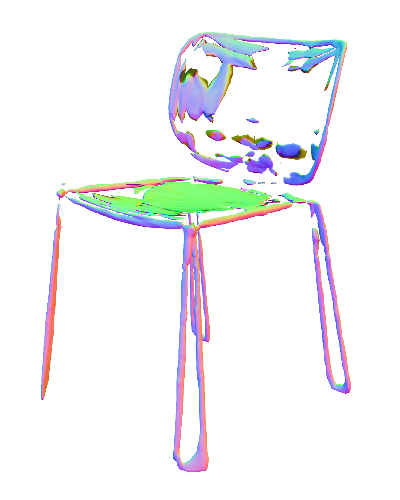}\vspace{-5pt}
        \caption*{Ours / LOD 1}
    \end{subfigure}%
    \begin{subfigure}[t]{0.2\linewidth}
        \centering
        \includegraphics[width=\textwidth]{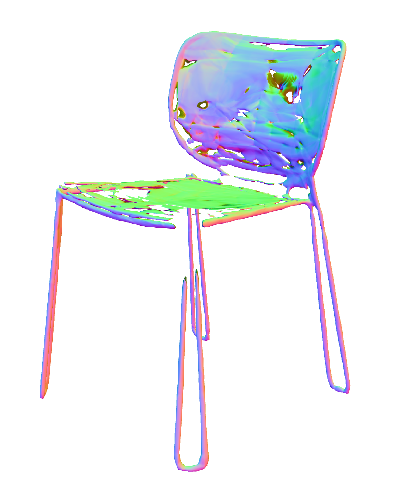}\vspace{-5pt}
        \caption*{Ours / LOD 2}
    \end{subfigure}%
    \begin{subfigure}[t]{0.2\linewidth}
        \centering
        \includegraphics[width=\textwidth]{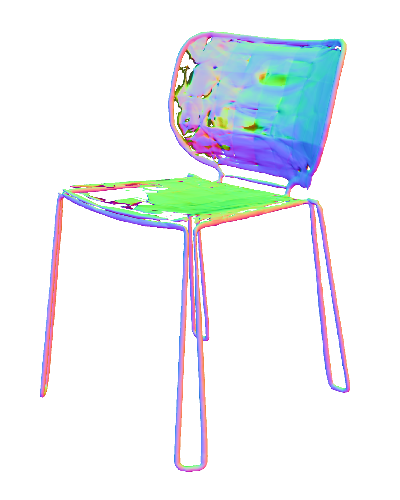}\vspace{-5pt}
        \caption*{Ours / LOD 3}
    \end{subfigure}%
    \begin{subfigure}[t]{0.2\linewidth}
        \centering
        \includegraphics[width=\textwidth]{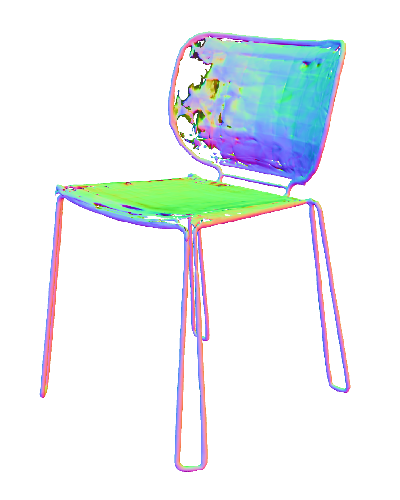}\vspace{-5pt}
        \caption*{Ours / LOD 4}
    \end{subfigure}%
    \begin{subfigure}[t]{0.2\linewidth}
        \centering
        \includegraphics[width=\textwidth]{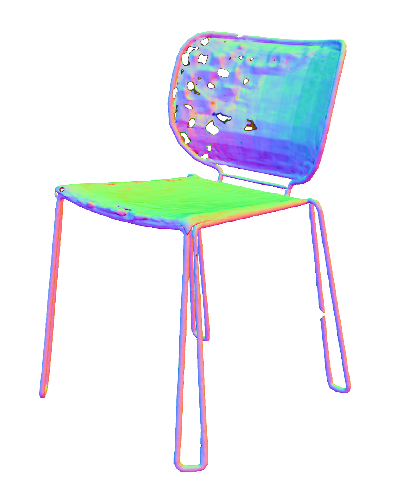}\vspace{-5pt}
        \caption*{Ours / LOD 5}
    \end{subfigure}%
    \caption{\textbf{Additional ShapeNet150 Results.} Our method struggles with 
    thin flat features with little to no volume,
    such as Jetfighter wings and the back of the Chair. Best viewed zoomed in.}
\label{fig:supp_sn_comp}
\vspace{-9pt}
\end{figure*}